\documentclass[Afour,sageh,times]{sagej}  
\usepackage{amsfonts}
\usepackage{amsmath}
\usepackage{amssymb}
\usepackage{amsthm}

\usepackage{algorithm}
\usepackage{algpseudocode}
\usepackage[UKenglish]{babel}
\usepackage{times}
\usepackage{graphicx}
\usepackage{epstopdf}
\usepackage{pdfpages}
\usepackage{multirow} 
\usepackage{multicol}
\usepackage{array}
\usepackage{mdwmath}
\usepackage{mdwtab}
\usepackage{rotating}
\usepackage{caption}

\usepackage{verbatim}
\usepackage{isomath}
\usepackage{overpic}
\usepackage[list=off]{caption}
\usepackage{standalone}
\usepackage{booktabs} 
\usepackage{siunitx} 
\usepackage{pgfplotstable} 
\usepackage{tikz} 
\usetikzlibrary{shapes,calc,patterns,
	decorations.pathmorphing,
	decorations.markings}

\usepackage[siunitx]{circuitikz} 

\usepackage{accents}
\newcommand{\ubar}[1]{\underaccent{\bar}{#1}}

\makeatletter\newcommand{\manuallabel}[2]{\def\@currentlabel{#2}\label{#1}}\makeatother

\usepackage{color}
\usepackage{xcolor}

\colorlet   {lightorange}{orange!20}
\colorlet   {lightgrey}  {gray!20}

\usepackage{xspace}
\newcommand{\newabbreviation}[2]{#1\ (\renewcommand{#1}{#2}#1)}

\newcommand{\RL     }{reinforcement learning}

\newcommand{\VIA    }{variable impedance actuator}

\newcommand{\VSA    }{variable stiffness actuator}

\newcommand{\VSAs}     {\VSA s}

\newcommand{\MACCEPAVD}{\MACCEPA\ with Variable Damping}
\newcommand{\MACCEPA}  {Mechanically Adjustable Compliance and Controllable Equilibrium Position Actuator}

\newcommand{\VIAs   }  {\VIA s}
\newcommand{\ILQR   }  {Iterative Linear Quadratic Regulator}
\newcommand{\SSM}{state-space model}
\newcommand{\OC}{optimal control}
\newcommand{\DMP}{dynamic motion primitive}
\newcommand{\DMPs}{\DMP s}
\newcommand{\IOC}{inverse optimal control}
\newcommand{\IRL}{inverse reinforcement learning}
\newcommand{\OCP}{optimal control problem}

\usepackage{amssymb}
\usepackage{mathtools}

\mathchardef\mhyphen="2D   

\newcommand{\RNum}[1]{\uppercase\expandafter{\romannumeral #1\relax}}

\newcommand{\defeq}{\vcentcolon=}

\newcommand{\T}     {\mathsf{T}}                

\newcommand{\intd}  {\mathrm{d}}          

\newcommand{\bO}    {\boldsymbol{0}}      
\newcommand{\bA}     {\mathbf{A}}         
\newcommand{\bC}     {\mathbf{C}}         
\newcommand{\bG}     {\mathbf{G}}         
\newcommand{\bH}     {\mathbf{H}}         
\newcommand{\bI}     {\mathbf{I}}          
\newcommand{\bJ}     {\mathbf{J}}         
\newcommand{\bK}     {\mathbf{K}}         
\newcommand{\bM}     {\mathbf{M}}         
\newcommand{\bN}     {\mathbf{N}}         

\newcommand{\bb}     {\mathbf{b}}         
\newcommand{\be}{\mathbf{e}}
\newcommand{\f}     {\mathbf{f}} 
\newcommand{\bg}{\mathbf{g}}        
\newcommand{\bp}{\mathbf{p}}
\newcommand{\bq}     {\mathbf{q}}         
\newcommand{\bu}  {\mathbf{u}}          
\newcommand{\bv}{\mathbf{v}}
\newcommand{\bt}{\mathbf{t}}

\newcommand{\bw}{\mathbf{w}}
\newcommand{\bx}     {\mathbf{x}}         
\newcommand{\bz}     {\mathbf{z}}

\newcommand{\btau}{\boldsymbol{\tau}}

\newcommand{\balpha}{\boldsymbol{\alpha}}
\newcommand{\bbeta}{\boldsymbol{\beta}}
\newcommand{\bepsilon}{\boldsymbol{\epsilon}}
\newcommand{\btheta}  {\boldsymbol{\theta}}               
\newcommand{\bxi}{\boldsymbol{\xi}}

\newcommand{\pinv}      [1]{{#1}^\dagger}

\newcommand{\qdot}   {\dot{q}}            
\newcommand{\bqdot}  {\dot{\bq}}          

\newcommand{\bqddot} {\ddot{\bq}}         
\newcommand{\bqdddot} {\dddot{\bq}}

\renewcommand{\btheta}     {\boldsymbol{\theta}} 
\newcommand  {\bthetadot}  {\dot{\btheta}}       

\newcommand  {\dt}         {\,\intd t}             
\newcommand  {\bpi} {\boldsymbol{\pi}}    

\newcommand{\inertia}{m} 

\newcommand{\Ein}{E_\mathrm{in}}

\graphicspath{
{figures/}
}

\providecommand{\figurename}{Fig.}
\newcommand*{\sref}[1]{\S\ref{s:#1}}            
\newcommand*{\tref}[1]{\tablename~\ref{t:#1}}   
\newcommand*{\fref}[1]{\figurename~\ref{f:#1}}  
\newcommand*{\eref}[1]{(\ref{e:#1})}            
\newcommand*{\cref}[1]{\chaptername~\ref{ch:#1}}

\usepackage[inline]{enumitem}
\setlist{nolistsep}
\newcommand{\il}[1]{\begin{enumerate*}[label=(\roman*)]#1\end{enumerate*}}

\newcommand{\eg}{\textit{e.g.,}~} %
\newcommand{\ie}{\textit{i.e.,}~} %
\newcommand{\etc}{\textit{etc.}}  %
\newcommand*{\taskref}[1]{Task \ref{task:#1}} 
\usepackage[normalem]{ulem}                                        
\usepackage{marginnote}
\usepackage[textwidth=10ex,colorinlistoftodos]{todonotes}

\colorlet{jb}{red}
\colorlet{mh}{red}
\colorlet{fw}{purple}

\renewcommand{\sout}[1]{}
\colorlet{change}{blue}      

\tikzstyle{spring}=[very thick,decorate,decoration={zigzag,pre length=2,post
	length=2,segment length=6}]

\tikzstyle{damper}=[thick,decoration={markings, 
	mark connection node=dmp,
	mark=at position 0.5 with 
	{
		\node (dmp) [thick,inner sep=0pt,transform shape,rotate=-90,minimum
		width=15pt,minimum height=3pt,draw=none] {};
		\draw [thick] ( $(dmp.north east)+(2pt,0)$ ) -- (dmp.south east) -- (dmp.south
		west) -- ( $(dmp.north west)+(2pt,0)$ );
		\draw [thick] ( $(dmp.north)+(0,-5pt)$ ) -- ( $(dmp.north)+(0,5pt)$ );
	}
}, decorate]

\usepackage{moreverb,url}
\newcommand\BibTeX{{\rmfamily B\kern-.05em \textsc{i\kern-.025em b}\kern-.08em
		T\kern-.1667em\lower.7ex\hbox{E}\kern-.125emX}}
\def\volumeyear{2020}

\begin{document}

\runninghead{Wu and Howard}

\title{Exploiting Variable Impedance for Energy Efficient Sequential Movements
}

\author{Fan Wu\affilnum{1} \affilnum{2} and Matthew Howard\affilnum{2}}
\affiliation{\affilnum{1}Technical University of Munich, Germany\\
	\affilnum{2}King's College London, UK}

\corrauth{Fan Wu, Munich School of Robotics and Machine Intelligence,
	Technical University of Munich, Munich,
	80797, Germany.}

\email{f.wu@tum.de}


\begin{abstract}
Compliant robotics have seen successful applications in energy efficient locomotion and cyclic manipulation. However, exploitation of variable physical impedance for energy efficient sequential movements has not been extensively addressed.
This work employs a hierarchical approach to encapsulate low-level optimal control for sub-movement generation into an outer loop of iterative policy improvement, thereby leveraging the benefits of both optimal control and reinforcement learning. The framework enables optimizing efficiency trade-off for minimal energy expenses in a model-free manner, by taking account of cost function weighting, variable impedance exploitation, and transition timing --- which are associated with the skill of compliance.
The effectiveness of the proposed method is evaluated using two consecutive reaching tasks on a variable impedance actuator. The results demonstrate significant energy saving by improving the skill of compliance, with an electrical consumption reduction of about $30\%$ measured in a physical robot experiment.
\end{abstract}


\maketitle

\section{Introduction}\label{s:introduction}
Intrinsically compliant robots typically have elastic components for stiffness modulation and such elements are capable of storing elastic energy. 
The field of robotic locomotion has seen a series of successful developments of energy efficient robots with elastic joints or springy legs that can exploit this energy storage.
It is of great interest to apply the same principle to robotic manipulators such that soft robots can behave in a human-like energy efficient way for a wide variety of tasks.

Biological springs, like tendons and various elastic elements in muscles, are embedded in humans and animals and make them highly efficient runners and jumpers (\cite{Roberts2016}).
Utilizing elastic energy storage and recoil, which is associated with optimizing muscular stiffness and transition timing, is a crucial skill that can be practised and improved for many other athletic activities, not limited to locomotion (\cite{Wilson2008}).

Physical compliance incorporating elastic components is prominent for energy efficient lower limb locomotion (\cite{Reher2016,Roozing2016DesignEfficiency,Roozing2019}).
Also, they have been demonstrated to reproduce the skill of \emph{energy buffering} in explosive movements such as throwing (\cite{Wolf2008,Braun2013}). Storing and discharging elastic energy, which was called as ``skill of compliance'' by Okada \cite{Okada2002}, can amplify the output power, exceeding the power limit of the drive motor.
Other recent studies attempt to improve energy efficiency for cyclic manipulation tasks, \eg repetitive pick-and-place (\cite{Matsusaka2016}) and dribbling a basketball (\cite{Haddadin2018}).


However, many tasks in unstructured environments are not periodic and 
\emph{variable physical impedance} is hard to fully exploit. 
For instance, the objects to be picked and placed may be located at random positions. The task given to a robot may consist of a sequence of different types of actions, such as ``reach a cup, grasp it, and pour the water". These non-periodic but \emph{sequential} tasks more commonly involve upper limbs and are complicated by their greater diversity.
The problem of task-oriented sequential movement generation --- in the context of compliant robotics --- faces the difficulty imposed by inherent actuation redundancy. The control redundancy of the actuators, which is somehow equivalent to the muscle redundancy of musculoskeletal arms, makes it non-trivial to optimize the movements in the ``muscle space''. 


\begin{figure*}[!htb]
	\centering
	\begin{overpic}[width=\textwidth]{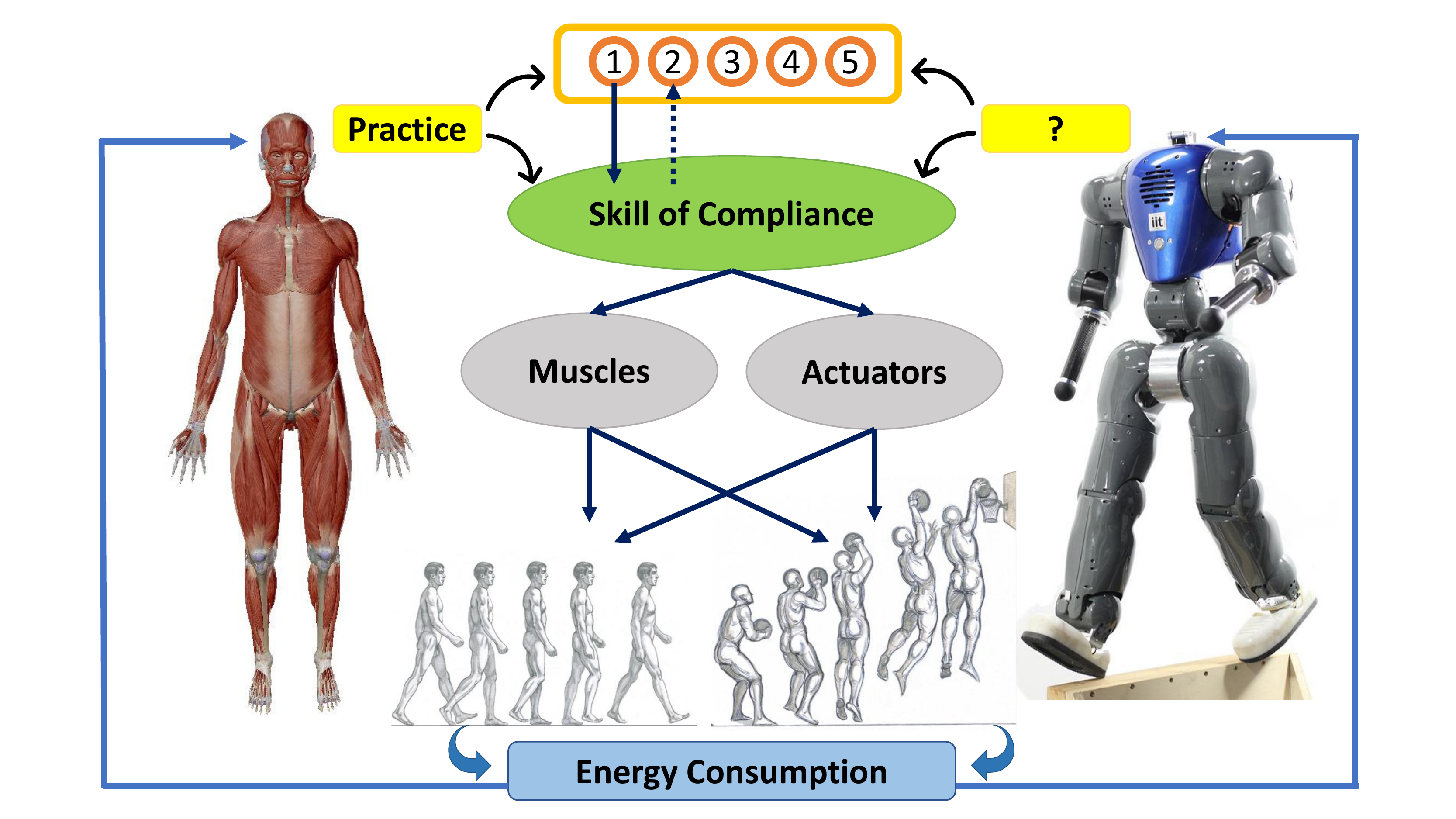}
		\put(15,5){\ref{f:seq_concept_human} Human}
		\put(75,5){\ref{f:seq_concept_robot} Robot}
	\end{overpic}
	\caption{\label{f:seq_concept} \textbf{Conceptual diagram of energy efficient sequential movements}.\endnote{The graphics showing the movements of a human walking and playing basketball are from the book \textit{Classic Human Anatomy in Motion: The Artist's Guide to the Dynamics of Figure Drawing} written by Valerie L. Winslow.} \begin{enumerate*}[label=(\Alph*)]
			\item \label{f:seq_concept_human}A human, and
			\item \label{f:seq_concept_robot}a compliant humanoid robot COMAN developed by \cite{Tsagarakis2013}.
		\end{enumerate*}
		A task is represented by a sequence of submovements illustrated by orange circles. When one of the submovements is triggered (circle "1" in the figure), humans can use the skill of compliance --- modulation of elastic energy storage and muscular stiffness --- to improve the performance and energy efficiency of subsequent actions (denoted by the dashed arrow pointing to circle "2"). The same strategy can be realized by robots with physical compliance. Humans can minimize the energy consumption of skilled movements via practice, resulting in improved movement representations and associated muscle skills. However, how physical compliance can be fully exploited to minimize energy cost for sequential movements still remains an under-explored question.}
\end{figure*}

Energetic economy is of great importance to reproduce human-like skilled movements. 
Researchers have embraced the notions of movement economy or efficiency since 1980s to understand and model human neuromuscular control of skilled movements (\cite{Nelson1983,Sparrow1998,Todorov2002}).
The emergence and learning of complex motor skills can be explained as an optimization process aiming at minimizing metabolic energy expenditure subject to task, environment and organism constraints.
Reduction of metabolic cost of human movements during training and practice has been verified by empirical studies (\cite{Lay2002,Huang2012}).
However, for robots driven by \newabbreviation{\VIAs}{VIAs}\renewcommand{\VIA}{VIA} which are viewed as the mechanical counterparts of humans and animals, there lacks a systematic optimal control approach to optimize energy efficiency of complex skills modelled as sequential movements (the analogy between humans and compliant robots in terms of energy efficient sequential movements is depicted in \fref{seq_concept}). 
To address this, our work postulates that such a framework should consider the following aspects:
\begin{enumerate}
	\item \textbf{Cost function weighting}. 
	Optimization of the weighting parameters of individual cost functions to achieve a higher level objective, \eg minimal energy consumption.\endnote{When the form of the cost function is determined, the weighting parameter can be adjusted to tune the energy efficiency. For simple quadratic control effort, the weight for each sub-movement need not be the same and can be optimized according to realistic energetics (by estimation or measurement).}
	\item  \textbf{Variable impedance exploitation}. A movement can adjust physical impedance (alongside the trajectory and at the transition phase) to improve its subsequent movements. 
	\item \textbf{Relative timing}. Temporal characteristics affect the energy efficiency. For instance, given a time horizon for the whole movement sequence, the relative timing of submovements is of importance for skilled efficient movements.
\end{enumerate}
These three issues have been addressed in part in the literature. For example, \IOC\ or \IRL\ is capable of learning the cost function from human demonstration (\cite{Mombaur2010,Berret2011,Levine2012}).
\cite{Nakanishi2016} exploited variable stiffness actuation for \emph{multiphase} movements by \OC, where a brachiation task is used for demonstration. 
\cite{Nakanishi2011} extended \newabbreviation{\OC}{OC} to include optimization of movement durations. An analogue via approximate inference was provided in \cite{Rawlik2010}.
Other works focus on optimizing the sub-goals or attractors of movements encoded by dynamical systems (\cite{Toussaint2007,Stulp2012}).

However, rarely have existing approaches addressed the above targets in the sequential context within one framework.
Also, many optimization-based methods rely on combining cost functions of subtasks into a composite one, which intensifies the cost function shaping issue --- requirement of redesigning the forms of cost functions --- when competing terms join together.

Therefore, this paper proposes a hierarchical approach that is capable of optimizing the three aspects identified above and mitigate the cost function shaping issue.
More specifically, a bi-level structure is employed to encapsulate a low-level \OC\ layer for submovement generation into an outer loop of iterative policy improvement, thereby benefits of both OC and RL are leveraged.
The high-level optimization formulated as a \RL\ problem enables optimizing the trade-off balance concerning (low-level) (1) cost function weighting, (2) variable impedance exploitation and (3) transition timing for minimal \emph{realistic} energetics.
The associated high-level policy parameters can be optimized in a derivative-free fashion by a black-box optimization (BBO) method for policy improvement suggested by \cite{Stulp2013}.
It can be viewed as a simplification of the RL algorithm $\mathrm{PI}^2$ (\cite{Theodorou2010JMLR}), which closely resembles an evolution strategy (ES) $(\mu,\lambda)-$ES, the backbone of CMA-ES algorithm (\cite{Hansen2001}).
At the low-level \OC\ naturally resolves the actuation redundancy and exploit variable impedance of \VIAs\ (\cite{Braun2012,Braun2013}), for which there exists efficient solvers \eg \ILQR\ (\cite{Li2004,Tassa2014}).

The rest of this paper is organized as follows. 
In \sref{ch6_related_work} we discuss relevant literature and concepts.
\sref{ch6_problem-def} first introduces a simple OC example of point-to-point reaching on the single joint \VIA. By investigating the \emph{efficient frontiers} of the OC problem we show how the hyper-parameters is identified and how the reinforcement learning problem is formulated. 
The proposed method is introduced in \sref{ch6_method}. 
Its effectiveness is evaluated by consecutive reaching tasks on a real \VIA\ robot. Simulations demonstrate significant energy efficiency improvement and a reduction of electrical consumption of about $30\%$ is recorded on the hardware.
Conclusions and future works are covered in \sref{ch6_conclusion}.

\section{Related Work}\label{s:ch6_related_work}

\subsection{Sequential Movements}
Sequential movements are common found in human daily life, from jaw movement for speech, finger movement for playing musical instruments, to many athletic whole body actions. 
How can these skilful human movements be learnt, executed and improved? 
Central to that is whether a hierarchical structure of representation, learning and control of movement sequences exists in the human brain.
The hypothesis of hierarchical organization of movement planning was proposed a long time ago in mid twentieth century by behaviourist Karl Lashley (\cite{Lashley1951}).
Recent experimental studies have provided evidence of hierarchical representation of movement sequences in the brain.
For instance, \cite{Yokoi2019} found that individual finger presses are represented in the primary motor cortex, whereas activities about the sequential context happen mainly in the premotor and parietal cortices. 

\begin{figure}[!htb]
	\centering
	\begin{overpic}[width=0.9\linewidth]{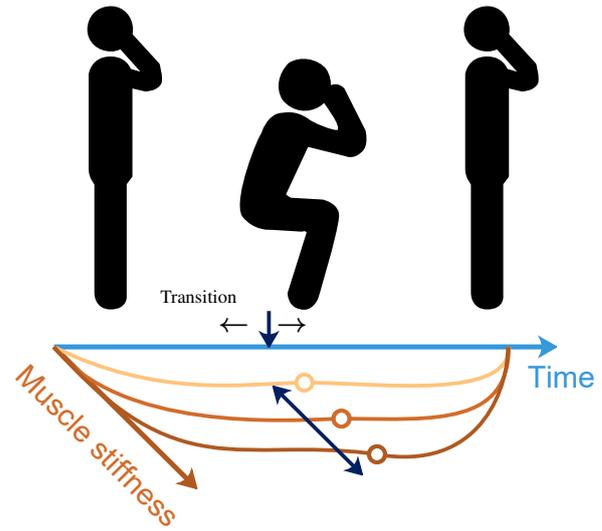}
		\put(25,40){\scriptsize Transition}
		\put(35,35){$\boldsymbol{\leftarrow} $}
		\put(45,35){$\boldsymbol{\rightarrow} $}
	\end{overpic}
	\caption{\label{f:squat}Humans can acquire new skilled movement by sequencing simpler motion primitives. A squat can be composed of crouching and rising-up, and the corresponding variables in the sequential context can be improved through practice. Possible ways to optimize the squat towards higher energy efficiency are: \il{\item adjust transition timing, and \item modulate muscular stiffness.} }
\end{figure}



In the robotics literature, sequential composition of controllers was employed by \cite{Burridge1999} to achieve dynamically dexterous robot behaviours. 
In robot learning control, motion generation of complex skills is often investigated at the task level and treated in a hierarchical manner.
A complex skill can be learnt from human demonstrations by motion segmentation into movement primitives (\cite{Lucia2013}).
Then a skilful movement can be composed by a ``repertoire'' (\cite{Schaal2010}) of such sequenced submovements.
By doing so it is expected to realize more general motion intelligence and make robots master interactive tasks and tool use, which is a hallmark of human behaviour (\cite{Hogan2012}).



Humans can acquire a new skilled movement by sequencing simpler motion primitives and improving via practice. 
Although each individual movement can be fine-tuned during training, the increased performance through practice can be clearly attributed to improvements in high-level planning processes, as shown by \cite{Ariani2019}.
For example, consider a squat (see \fref{squat}) that can be composed of crouching and rising-up.
By intuition, the contextual variables at the sequence planning level can possibly be transition timing, muscular stiffness, torque distribution, \etc\
\cite{Motegi2011} used \OC\ to find optimal transition timing that can reproduce experimentally measured human squat movements. 
The role of stiffness was investigated by \cite{Bobbert2001} also through biomechanical modelling and OC, which signifies the importance of \emph{exploiting elastic energy storage}.



\subsection{Optimization of Sequential Movements}
Improvement of a sequential movement necessitates existence of redundancy in either representational level or control level. In the above squat example, the transition timing is not predefined by the task or sub-movements, and thus can be tuned. While for playing a piece of music, the tempo and rhythm are determined, then the transition timing is specified by the task objective and cannot be exploited.   
\begin{figure}[!htb]
	\centering
	\begin{tikzpicture}
	\node[anchor = south west,inner sep=0] (image) at (0,0) {\includegraphics[width=\linewidth]{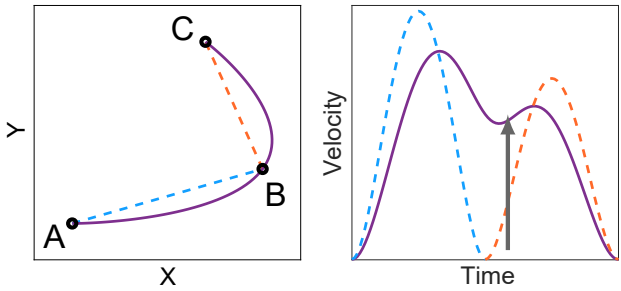}};
	\draw[-latex, black!60, ultra thick] (6.6,0.5) -- (6.6,2.3); 
	\end{tikzpicture}
	\caption{\label{f:ch6_concept_minjerk}Minimal jerk trajectories AB, BC and a via-point movement AC. AB, BC are both individually minimal jerk trajectories, but simply sequencing them is not optimal for A via B to C. The optimal via-point minimal jerk trajectory is curved around B in the X-Y plane (right). }
\end{figure}

It is easy to notice that sequentially combining the sub-movements, which are optimized with respect to their sub-goals, does not necessarily result in the optimal movement for the whole task. 
Look at the example illustrated in \fref{ch6_concept_minjerk}, the minimal-jerk trajectory (\cite{Flash1985})
of a via-point task from point A to C via B (at a specific time) results in a curved path in the X-Y plane.
While the minimal-jerk model of a single point-to-point movement always shows a straight path.
Consequently, if we sequence AB and BC (both individually are minimal-jerk) directly, the resulting trajectory is not optimal in terms of the whole movement. 
The difference is simply due to that the velocity at the via point is not constrained to be zero. 
Though it is obvious, this common phenomenon in kinematic domain shows an example of exploiting the \emph{redundancy} of velocity profile when concatenating discrete movements.

\begin{figure}[!t]
	\centering
	\begin{tikzpicture}[mass/.style={draw,thick}]
		\begin{scope}[scale=0.70]
			\node at (0,0) {\RNum{3}:}; 
			\node[anchor=west,mass,dashed] at (1,0) (type4pi1) { $\bpi_1 \defeq \arg\min  ~J_1 $ };
			\node[anchor=west,mass,dashed] at (1,-1.2) (type4pi2) { $\bpi_2 \defeq \arg\min ~J_2 $ };
			\node[anchor=west,mass,dashed] at (1,-2.4) (type4pi3) { $\bpi_3 \defeq \arg\min ~J_3 $ };
			\node[anchor=west] at (1,-4) (type4pi4) { $\displaystyle \bxi \leftarrow \bxi + \frac{\partial J_{\mathrm{total}}}{\partial \bxi}  $ };
			\draw[-Latex] (type4pi1.south) -- (type4pi2.north);
			\draw[-Latex] (type4pi2.south) -- (type4pi3.north);
			\draw[-Latex] (type4pi3.south) -- (type4pi4.north-|type4pi3);
			\draw[-Latex] (type4pi4.west) --++(-0.5,0) |- (type4pi1.west);
		\end{scope}
		
		\begin{scope}[scale=0.7,shift={(-7,-2)}]
			\node at (0,0) {\RNum{2}:}; 
			\node[anchor=west,mass,dashed] at (1,0) (type4pi1) { $ \{\bpi_i\} \defeq \arg\min ~ \sum J_i $ };
			\node[anchor=west] at (1,-1.6) (type4pi4) { $\displaystyle \bxi \leftarrow \bxi + \frac{\partial J_{\mathrm{total}}}{\partial \bxi} $ };
			\draw[-Latex] (type4pi4.west) --++(-0.5,0) |- (type4pi1.west);
			\draw[-Latex] (type4pi1.south) -- (type4pi4.north-|type4pi1);
			\node[anchor=west] at (0,-2.6) {\small $Examples: \textnormal{ILQR-T}, \textnormal{AICO-T}$};
		\end{scope}

		\begin{scope}[scale=0.7,shift={(-7,0)}]
			\node at (0,0) {\RNum{1}:}; 
			\node[anchor=west,mass,dashed] at (0.5,-0.2) (type4pi1) { $  \{\{\bpi_i \} , \boldsymbol{\xi}  \} \defeq \underset{ \{ \boldsymbol{\pi}_i \},\boldsymbol{\xi} }{\arg\min} ~ \sum J_i $ };
			\node[anchor=west] at (0,-1.1) {\small $Example: \mathrm{PI^2SEQ}$} ;
		\end{scope}
		
		\node at (-1.0,-3.8) {$(\textnormal{a})$};
	\end{tikzpicture}
	
	\small
	\begin{tabular}{ |c||c|c|c| }
		\hline
		& \RNum{1} & \RNum{2} & \RNum{3}  \\
		\hline
		Optimize cost function weighting & No   & Yes & Yes \\
		\hline
		Exploit variable impedance & Yes  & Yes & Yes \\
		\hline
		Optimize temporal parameters & Yes  & Yes & Yes \\ \hline
		Avoid redesign of composite cost function & No& No& Yes\\
		\hline
	\end{tabular}\vfill
	\vspace{2mm}
	(b)
	\caption{\label{f:ch6_concept_types}\begin{enumerate*}[label=(\alph*)]
			\item Three possible types of approaches for sequential movement optimization.  Dashed rectangle means a full optimization loop.
			\item \label{f:ch6_concept_types_table}The table summarizes comparison of type \RNum{1}-\RNum{3}. For simplicity, in Type \RNum{3} only 3 sub-problems are shown to visualize a sequence.
	\end{enumerate*}}
\end{figure}

Based on the above reasoning, it follows that when a sequence is generated by chaining movement primitives, it may be suboptimal without appropriately planning each individual considering the whole trajectory or its subsequent ones. In general, one can structure the problem as a composite optimization or tackle it hierarchically. Depending on whether either way is adopted, or both, there are three possible approaches, as depicted in \fref{ch6_concept_types}. 
Throughout this paper, $\bpi$ denotes the control policy, $\bxi$ represents vector of policy parameters to be optimized via reinforcement learning, $J$ is used for cost function.

To narrow down our discussion, we mainly consider the applications for \il{\item optimization of cost function weighting, \item exploitation of variable impedance, and \item optimising temporal parameters such as the time horizon and relative timing.}


To avoid confusion, the sequential movements/tasks considered in this chapter
are sequences in a predefined order. The problem of planning the order of executing a set of actions for a given task is not within the scope of this work. This kind of task planning problem does not predefine an order of executing subtasks.
Therefore it needs a higher level planning to figure out the best order to chain the submovements, typically from a discrete set of actions (\cite{Manschitz2015}).

\subsubsection{Composite Optimization}
Composite optimization here means optimizing w.r.t. a composite cost function that consists of the objectives of subtasks. 
For instance, an optimization-based approach usually consider the via-point problem by defining the cost function as 
\begin{align}
J = &(\bx(t_v) - \bx^*_v )^\T \bH_v (\bx - \bx^*_v)\nonumber\\ 
 &+ (\bx(t_f) - \bx^*_f )^\T \bH_f (\bx - \bx^*_f) \label{eq:J_concept_whole}
\end{align}
Here $\bx$ is the state vector of the problem, $\bx_v^*,\bx_f^*$ are the via-point and final targets respectively, $\bH_v,\bH_f$ are diagonal matrices to penalize the deviation, and $t_v,t_f$ represent the fixed via-point time and final time. 
The optimal control $\bu(\bx,t)=\bpi(\bx,t)$ with corresponding policy $\bpi$ is the one that minimizes the cost function.
The above minimal jerk via-point problem is one example that has analytical solution (\cite{Flash1985}).
The shortcoming of this is that if $t_v$ is allowed to be adjusted, the optimization of \eqref{eq:J_concept_whole} with a guess about $t_v$ may leads to suboptimal solutions.

Let us first consider the possibility to simultaneously optimize the control and some hyper-parameter like $t_v$. This is categorized as Type \RNum{1} in \fref{ch6_concept_types}.  
For many non-linear real problems arising in robotics, a classical method is to convert the \OC\ problem into a non-linear programming problem.
Considering the computational efficiency, a more efficient paradigm for learning control is to transform the representation of the control policies into a lower-dimensional space, and then optimize the policies and their hyper-parameters simultaneously. For example, \cite{Stulp2012} implemented the (model-free) reinforcement learning algorithm $\mathrm{PI^2}$ for sequential tasks (termed as $\mathrm{PI^2SEQ}$), with the help of \DMPs \renewcommand{\DMP}{DMP} (\DMP) for trajectory encoding using dynamical systems.
The shape parameter of trajectories and the attractors of dynamical systems are optimized together, so that the trajectory and its final state is optimized for all subsequent actions. The limitation of composite cost function is that it faces the cost function shaping issue. 
When competing terms from different subtasks come together, optimality of sub-movements may be no longer achievable.
In order to achieve optimality for all subtasks the formulation of the cost functions have to be redesigned.

\subsubsection{Hierarchical Optimization}
The second possible approach is to construct the optimization problem hierarchically. As shown by Type \RNum{2} in \fref{ch6_concept_types}, it is hierarchical in the sense that an inner loop and an outer loop optimize the control policies and hyper-parameters separately. 
Various previous studies addressing the multiphase optimal control can be found in this type.
To name a few, temporal optimization with \ILQR\ (ILQR-T) and approximate inference (AICO-T) was proposed by \cite{Nakanishi2011} and \cite{Rawlik2010} respectively. 
\cite{Nakanishi2011} used finite difference to compute the gradient of total cost w.r.t. change of time durations. 
The evaluation of the gradient is based on running the time-scaled augmented control and hence is very efficient. This is done by leveraging a technique that maps the real time to a canonical time.
It was demonstrated by \cite{Rawlik2010} with similar technique on a via-point task, where the algorithm finds an optimal relative timing. 
In case $J_{\mathrm{total}}$ is non-differentiable w.r.t. $\bxi$, one can utilize derivative-free methods (\cite{Conn2009}) such as trust region technique (\cite{Yuan2015}) and evolutionary strategy (\cite{Hansen2001}) in the outer loop.

This hierarchical structure coincides with the so-called ``bi-level'' problem in \IOC\ (\cite{Mombaur2010}).
In \IOC\ the outer loop optimizes the cost function shaping to match data demonstrated from a human.
Of interest here is the fact that the objective $J_\mathrm{total}$ in the outer loop need not be the same as the composite cost. Suppose that, for the speed and robustness of optimization, the subtasks may be described with simple quadratic terms such as traditional ``control effort'', or even have different energetic functions individually, but on the high level, the parameter can be updated according to more realistic cost estimator or physical measurement. This potential can be realized within the bi-level architecture.

Note that, since Type \RNum{2} also employ a composite cost function in the inner loop, it shares the same shortcoming with Type \RNum{1} that composite optimization may fail to achieve optimality for all subtasks and thus need redesign.
To overcome the drawback, we propose to optimize the sub-movements according to their own cost function as well as integrate the hierarchical (bi-level) architecture, which leads to Type \RNum{3} (\fref{ch6_concept_types}). The comparison against previous two types is summarized in the table (\fref{ch6_concept_types}~\ref{f:ch6_concept_types_table}).






\section{Problem Formulation}\label{s:ch6_problem-def}
In this section we first present a \OC\ model of a single joint driven by a \VIA, followed by an investigation of energy efficiency based on the concept of efficient frontiers. The intuition gained thereby helps with justifying the problem formulation.
Finally, a reinforcement learning problem is formulated that enables optimizing high-level parameters using policy improvement methods.

\subsection{A Simple Reaching Movement Model}\label{s:ch6_problemdef_reaching} 
Consider a point-to-point fast reaching task using a single-link robot driven by a \VIA.
The robot used in this paper is MACCEPA\renewcommand{\MACCEPA}{MACCEPA} \cite{VanHam2007} with variable damping \cite{Radulescu2012} (VD)\renewcommand{\MACCEPAVD}{MACCEPA-VD}. 
As illustrated in \fref{maccepavd}, the equilibrium position (EP) is controlled with SERVO1 and stiffness is regulated by spring pretension via SERVO2.
The mechanism was implemented in our previous work \cite{Wu2020},
where the damping is modulated by controlling a dedicated switching circuit that adjusts back-electromotive force on a DC motor attached to the joint. 
The system model provided in Appendix A.
\begin{figure}[!htb]
	\centering
	\includegraphics[width=0.95\linewidth]{drawing/maccepavd.tex}
	\caption{Diagram of MACCEPA-VD (\cite{VanHam2007,Radulescu2012}). 
	}
	\label{f:maccepavd}
\end{figure}
A fast reaching task is represented by a cost function
\begin{align}
J(x(\cdot),u(\cdot)) &= H(x(t_f)) + \int_0^{t_f} l(x(t),u(t),t) \dt \label{eq:ch6_fastreach_example_J}\\
H(x(t_f)) &= 1000(q(t_f)-q^*)^2 \label{eq:ch6_fastreach_example_H}\\
l(x(t),u(t),t) &= 1000 (q(t)-q^*)^2 + w_e( ( u_1(t)-q^* )^2 \nonumber\\
& \quad + u_2^2(t) + 10^{-3} (u_3(t)-0.5) ) \label{eq:ch6_fastreach_example_l}
\end{align}
An optimal control problem can be formulated as to seek an optimal control $\bu(t) \in U \in \mathbb{R}^3$ constrained by its admissible set $U=\{ \bu \in \mathbb{R}^3 \, | \, \bu_{\mathrm{min}} \preceq \bu \preceq \bu_{\mathrm{max}} \}$, that minimizes the cost function \eqref{eq:ch6_fastreach_example_J} and subject to the \SSM\ of the robot dynamics. In the cost function, $w_e$ serves as a weighting parameter to enable adjustment of the performance-cost trade-off.
  


In addition to trade-off balance via cost function weighting, the stiffness at transition could have a significant influence on the energy efficiency of the subsequent movement. This is explained as follows.

\subsection{Efficient Frontiers of Optimal Control}\label{s:ch6_efficient_frontiers}\noindent
\begin{figure*}[!htb]
	\centering
	\begin{overpic}[width=0.9\linewidth]{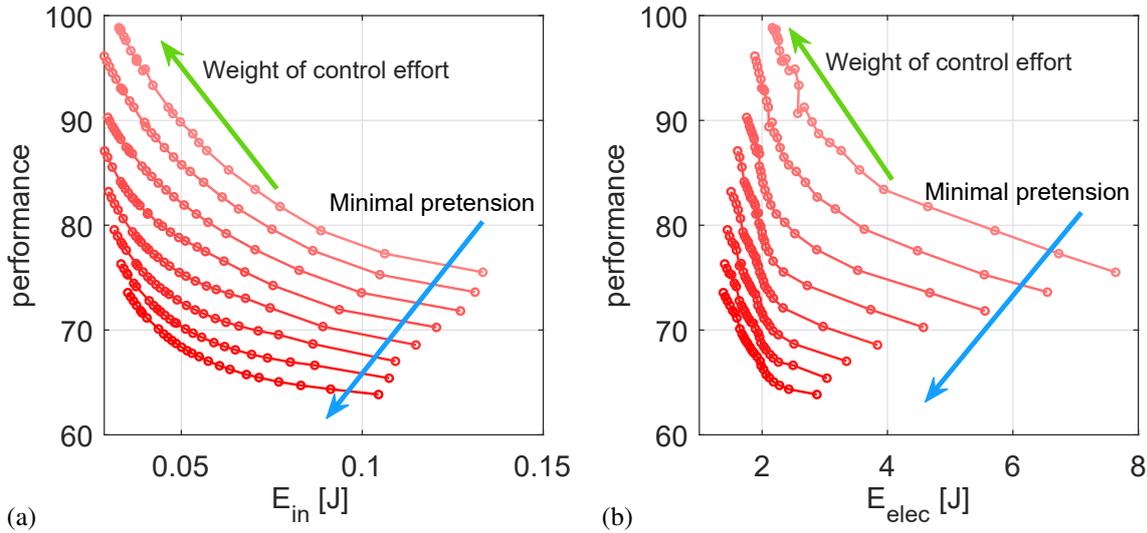}
		\put(0,0){\ref{f:pareto-Ein}}
		\put(50,0){\ref{f:pareto-Eelec}}
	\end{overpic}
	\caption{\label{f:pareto} Efficient frontiers :
		\begin{enumerate*}[label=(\alph*)]
			\item\label{f:pareto-Ein} fast reaching performance against input mechanical work, and
			\item\label{f:pareto-Eelec} fast reaching performance against electrical work.
		\end{enumerate*}
		Each efficient frontier shows the optimal control solutions by varying the weight $w_e$ of control effort term (shown by the green arrow) in the cost function, with a certain minimal spring pretension. 
		The green arrow indicates the direction of increasing the weight.
		The spring preset parameter $p_s$ is adjusted by the servo $\mathrm{M}_2$ from $0.1\,\mathrm{rad}$ to $1.5\,\mathrm{rad}$ with increment of $0.2\,\mathrm{rad}$. Increasing the minimal spring pretension (shown by the blue arrow) moves the efficient frontier downward, which means a increased overall energy efficiency. 
	}
\end{figure*}

Efficient fronter (EF) is a common tool to examine the trade-off of two competing objectives in an optimization problem. 
In this work the problem can be interpreted as optimizing the task performance while minimizing the energy cost.
The EF is then the set of optimal solutions that achieve the best performance at a defined energy cost.
The above OC problem has an efficient frontier by varying the weighting parameter $w_e$. Then the distribution of optimal solutions can be visualized in performance-cost plane.

In addition, to investigate how pre-stored elastic energy affect the energy efficiency, we generate the optimal solutions by \ILQR\ for different values of $w_e$, with a certain minimal spring pretension to produce an EF. Multiple EFs are generated by changing the condition of minimal spring pretension. This is done by setting the initial stiffness motor angle $\theta_2(0)$ and the lower bound $u^{(2)}_{\mathrm{min}}$ of $u_2$ to a preset value $p_s$, \ie let $\theta_2(0)=u^{\mathrm{min}}_{2}=p_s \in P_s \defeq \{ p_s \in \mathbb{R} ~|~ \theta_2^{\mathrm{min}} \leq p_s \leq \theta_2^{\mathrm{max}} \} $.

The results are shown in \fref{pareto}. 
The vertical axis represents the reaching accuracy performance, which is the terminal cost \eqref{eq:ch6_fastreach_example_H} plus the integral of the first term of running cost \eqref{eq:ch6_fastreach_example_l}. The horizontal axis is the energy cost, measured by positive input mechanical work\endnote{We assume that the motors are not back-drivable, thus no negative mechanical work to the motors can be regenerated. Similarly, the electrical energy is defined as the integral of the positive part.}  $E_{\mathrm{in}}$ and electric work $E_{\mathrm{elec}}$, both estimated by simulation.\endnote{Note that, the accuracy of estimating $E_{\mathrm{in}},E_{\mathrm{elec}}$ is very sensitive to simulation step size. For \ILQR\ we typically use time step $\Delta t=0.02$. While computing $E_{\mathrm{in}},E_{\mathrm{elec}}$ is based on simulation (of forward dynamics) with $\Delta t=0.001$.}

\begin{align}
E_{\mathrm{in}} = \int [P_{\mathrm{in1}}]^+ + [P_{\mathrm{in2}}]^+ \dt \label{e:Ein} \\
E_{\mathrm{elec}} = \int [P_{\mathrm{elec1}}]^+ + [P_{\mathrm{elec2}}]^+ \dt
\end{align}
where $[\cdot]^+=\max(0,\cdot)$. For in-depth analysis of modelling motor energy consumption, we refer the readers to \cite{Verstraten2016}. Calculation of the mechanical and electrical power is given in Appendix A. 

Looking at \fref{pareto}, when increasing the weight of control effort $w_e$ (as shown by the direction of green arrow), both mechanical and electrical consumption are decreased, with some loss of reaching performance. It demonstrates that even with a simple quadratic control cost, it is still possible to tune the trade-off between performance and realistic energy measures.
Moreover, it can be seen in that, by increasing the minimal spring pre-tension $p_s$, the efficient frontiers move towards the bottom-left, which signals an overall improvement of energy efficiency. 

Overall, the above investigation based on the tool of efficient frontier suggests that control cost weight $w_e$ and minimal stiffness $p_s$ can be taken as hyper-parameters that tunes the performance-cost trade-off according to realistic energy measures.

\subsection{Reinforcement Learning Formulation}\noindent
Based on the previous rapid reaching \OC\ model, let us now consider a consecutive reaching task for example, that requires the arm to reach a sequence of targets $\{q^*_i\}_{i=1}^{N_s} $ from initial state $\bx_0$, where $N_s$ is the number of subtasks. The sequential movement $\mathcal{S} \coloneqq \{\mathcal{M}_i \}_{i=1}^{N_s} $ consists of $N_s$ sub-movements generated by solving \OCP\ (\renewcommand{\OCP}{OCP}\OCP).
The sub-problems is denoted as $\{\mathrm{\OCP}_i \}_{i=1}^{N_s}$.
We define $\bxi \in \{ \bxi \in \mathbb{R}^{d_p} ~|~ \bxi_\mathrm{min} \preceq \bxi \preceq \bxi_\mathrm{max} \}$ to be the stacked vector of weighting parameter $\bw_e=\{w^{(i)}_e \}$, stiffness parameter $\bp_s = \{ p^{(i)}_s \}$, and movement durations $\bt_d= \{ t_d^{(i)} \}$.\endnote{By convention, all vector quantities are assumed to be column vectors.} 
$\bxi_\mathrm{min},\bxi_\mathrm{max} $ are lower and upper bound of $\bxi$.
Note that, depending on the type of task at hand, $\bp_s$ may have different meaning. For example, as in \sref{ch6_efficient_frontiers} it is used to set the minimal stiffness motor command. By doing so it constrains the minimal elastic energy to be stored and sets a target for the motor.


Our problem is to find an energy optimal trajectory $\mathcal{S}$ and $\bxi$ that minimizes energy cost while achieving all sub-goals. Mathematically, it is formulated as to minimize the episodic cost:
\begin{align}
J(\mathcal{S}) = J_e +  \mathcal{C} \cdot \max\{0, J_p - \bar{J}_p\} \label{e:Jmod}
\end{align}
The cost objective \eref{Jmod} is formulated as an \emph{episodic} cost. 
$J_e$ is the energy consumption, and $J_p$ is the cost associated with task achievement. The amount of $J_p$ exceeding an upper bound $\bar{J}_p$ is penalized by a large constant $\mathcal{C}$. The energy consumption can be estimated by a cost function or measured on hardware. $\bar{J}_p$ is evaluated by solving $\{ \mathrm{OCP}_i \}$ with initial $\bxi^{(0)}$.

\section{Method}\label{s:ch6_method}
The policy improvement optimizes $J$ in an iterative process. \fref{ch6_method_diagram} outlines the paradigm of our proposed policy improvement method which encapsulates OCPs at the low-level. It consists of the main steps of general policy improvement procedures: exploration, evaluation, and policy update.
Different from the vanilla \RL\, from exploration and evaluation we have an inner loop to solve $\{ \mathrm{\OCP}_i \}$ sequentially.

\begin{figure}[!ht]
	\centering
	\begin{tikzpicture}
	\node[anchor=south west] at (0,0) {\includegraphics[width=\linewidth]{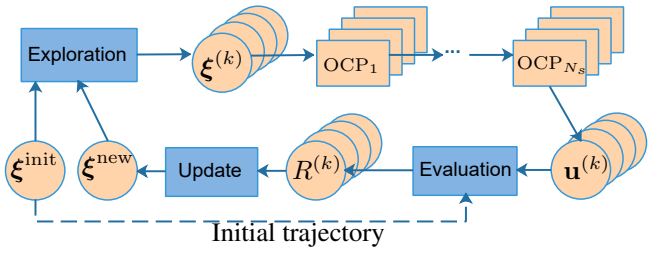}};
	\node at (0.53,0.9) {$\bxi^{\mathrm{init}}$};
	\node at (1.5,0.9) {$\bxi^{\mathrm{new}}$};
	\node at (4.25,0.9) {$R^{(k)}$};
	\node at (7.8,0.9) {$ { \bu^{(k)} } $};
	\node at (7.35, 2.25) {\scriptsize $\mathrm{OCP}_{N_s}$};
	\node at (4.7, 2.25) {\scriptsize $\mathrm{OCP}_{1}$};
	\node at (3., 2.25) {$\bxi^{(k)}$};
	\node at (4.,0) {Initial trajectory};
	\end{tikzpicture}
	\caption{\label{f:ch6_method_diagram} Diagram of proposed policy improvement method.  }
\end{figure}

The first step is to evaluate the initial trajectory with $\bxi^{(0)}$ given by the user. Once $\{ \mathrm{\OCP}_i \}$ are specified, we run \ILQR\ to generate $\mathcal{S}^{(0)} = \{ \mathcal{M}_i \}_{i=1}^{N_s} $ and obtain corresponding costs $J_e^{(0)},J_p^{(0)}$.\endnote{For the initial trajectory, it is obvious that $J^{(0)} = J_e^{(0)}$}
The task performance constraint is set up by multiplying a tolerance factor $\sigma_{\mathrm{tol}} \in \{ 0 \cup \mathbb{R}^+ \}$ with $J_p^{(0)}$, \ie $\bar{J}_p = (1+ \sigma_{\mathrm{tol}} )\, J_p^{(0)} $.
The tolerance factor is introduced for user to trade-off the energy efficiency flexibly. 
A positive value allows the exploration for some samples that have worse performance so that the information may contribute to faster and more robust updating towards the minimal energy cost.


\subsection{Exploration and evaluation}
The exploration phase generates $K$ unconstrained perturbations in policy parameter space for $K$ roll-outs. The perturbations $\tilde{\epsilon}_k \sim \mathcal{N}(\mathbf{0}, \gamma^{n-1} \boldsymbol{\Sigma} _\epsilon), (k=1,...,K$ is assumed to obey normal distribution, where $\boldsymbol{\Sigma}_\epsilon$ is the covariance matrix and $\gamma \in (0,1)$ is the decay factor. Then the box constraint $\mathbf{\ubar{b}}$ and $\mathbf{\bar{b}}$ is applied to yield
\begin{align}
\bepsilon_k &= \min(\max(\tilde{\bepsilon}_k+\bxi^{(n)},\bxi_\mathrm{min}),\bxi_\mathrm{max})-\bxi^{(n)} \\
\bxi^{(n)[k]} &= \bxi^{(n)} + \bepsilon_k
\end{align}
When running each $k$-th roll-out, $\bxi^{(n)[k]}$ is used to specify sub-problems with $\bw_e$ for cost function $J_i$, $\bp_s$ for stiffness motor constraint, and $\bt_d$ for time horizon. 
Without loss of generality, we assume the time horizon $[t_0,t_f]$ of $i$-th sub-problem is from $t_0=0$ to $t_f=t_d^{(i)}$.

With these details, $\{ \mathrm{\OCP}_i \} $ are solved by \ILQR\ sequentially to generate the sub-movements $\mathcal{M}_i$. The final state of $\mathcal{M}_i = \{ \bx_i, \bu_i  \} $ is taken as the initial state of its sequent problem, \ie $ \bx_{i+1}(0) = \bx_{i}(t^{(0)}_f)$.
The energy consumption $J_e^{[k]}$ and task performance $J_p^{[k]}$ along the trajectory is then evaluated by running a forward pass of dynamics and control $\bu = \{ \bu_i \}$.
After running K roll-outs and collecting relevant costs, the total costs $J^{[k]}$ are calculated by \eref{Jmod}.

\begin{algorithm}[t]
	\caption{Optimization of sequential movements using OC-ES}\label{algo_OCES}
	\begin{algorithmic}[1]
		\State \textbf{Given}: $\{J_i\}$, $\{\mathrm{\OCP}_i\}$, $\bxi_\mathrm{min},\bxi_\mathrm{max}$
		\State \textbf{Initialization}: $\bxi^{(0)}, \gamma,\boldsymbol{\Sigma_\epsilon},\mu,\sigma_{\mathrm{tol}} $
		\State Generate $\mathcal{S}^{(0)}$ by solving $\{\mathrm{\OCP}_i\}$, compute $\bar{J}$
		\Repeat
		\For{$k = 1$ to $K$}
		\Comment{k-th rollout}
		\State Sample $\tilde{\bepsilon}_k \sim \mathcal{N}(\mathbf{0},\gamma^{n-1} \boldsymbol{\Sigma}_\epsilon)$ 
		\Comment{Unconstrained perturbations}
		\State $\bepsilon_k = \min(\max(\tilde{\bepsilon}_k+\bxi^{(n)},\bxi_\mathrm{min}),\bxi_\mathrm{max})-\bxi^{(n)}$
		\Comment{Constrained perturbations}
		\State $\bxi^{(n)[k]} \leftarrow \bxi^{(n)}+\bepsilon_k$
		\State Specify hyper-parameters and constraints of $\{\mathrm{\OCP}_i\}$ according to $\bxi^{(n)[k]}$
		\For{$i=1$ to $N_s$}
		\State $t_0=0,t_f=t_d^{(i)}, \bx_i(0)=\bx_{i-1}(t_f )$, 
		\State solve $\{\mathrm{\OCP}_i\}$, $\bu_i = \arg\min J_i  $
		\State  $\mathcal{M}_i=\{ \bx_i, \bu_i \}$ 
		\EndFor
		\State Estimate $J_p^{[k]},J_e^{[k]}$, 
		\EndFor
		\State Retrieve stored samples and append to dataset $\{ J^{[k]},\bepsilon_k \}$, $K'=K+\mu$
		\State Compute and normalize cost $\{J^{[k]} \}_{k=1}^{K'}$ by \eref{Jmod} \eqref{eq:ch6_update_normal}
		\State Update $\xi^{(n+1)}$ using \eqref{eq:ch6_update_prop} and \eqref{eq:ch6_update_update}
		\State Keep $\mu$ best samples for sample reuse
		\Until $\bxi$ converges or maximum number of iterations reached
	\end{algorithmic}
\end{algorithm}

\subsection{High-level policy update}
The policy update step \eqref{eq:ch6_update_normal}-\eqref{eq:ch6_update_update} utilizes the reward-weighted averaging rule as introduced by \cite{Stulp2013}.
\begin{align}
\tilde{J}^{[k]} &= \frac{J^{[k]} - \min( \{ J^{[k]} \} )}{\max(\{ J^{[k]} \}) - \min( \{ J^{[k]} \} )} \label{eq:ch6_update_normal} \\
P_k & = \frac{ \exp(-c \tilde{J}^{[k]}) }{\sum_{i=1}^{K} \exp(-c \tilde{J}^{[i]}  ) } \label{eq:ch6_update_prop} \\
\bxi & \leftarrow \bxi + \sum_{k=1}^{K} P_k \bepsilon_k \label{eq:ch6_update_update}
\end{align}
First the cost $J^{[k]}$ is normalized according to their maximum and minimum by \eqref{eq:ch6_update_normal}. The normalized cost $\tilde{J}^{[k]}$ is used to calculate probability $P_k$ for $k$-th roll-out according to \eqref{eq:ch6_update_prop}, where $c>0$ is a constant.\endnote{In our implementation we choose $c=10$.}
Finally, the update is computed by the weighted averaging rule \eqref{eq:ch6_update_update}.

The above weighted averaging technique is simplified from $\mathrm{PI}^2$ (\cite{Stulp2013}) and 
converts the policy improvement method into a black-box
optimization (BBO) method that resembles the evolutionary strategy $(\mu,\lambda)-$ES, which is the basic form of CMA-ES algorithm (\cite{Hansen2001}).
It is appealing because it can solve non-linear non-convex black box optimization problems with reasonable efficiency. 
Unlike CMA-ES (\cite{Hansen2001}), it does not have the covariance matrix adaption step. Instead, we manually specify a decay factor $\gamma$ to gradually decrease the variance of perturbations. 

A $(\mu,\lambda)-$ES method consists of three steps: mutation, selection and recombination.
The exploration phase corresponds to the mutation step. Then all samples
are selected for policy update (recombination). The policy update step can be
viewed as recombination of samples. In contrast to the reinforcement learning
algorithm $\mathrm{PI}^2$ that leverages the problem structure, ES treats the policy improvement as a BBO problem. Since the high-level optimization is solved as a BBO problem in our proposed policy improvement method, we label the high-level part of the whole method as a evolutionary strategy.

For better robustness of convergence, another technique employed is sample reuse. After every update, we keep $\mu$ best samples among $K$ roll-outs (at current iteration) for next update. Therefore, after the first iteration, we have $K+\mu$ samples. 
The exploration, evaluation and policy update procedures are repeated until $\bxi$ converges or reaches maximum steps.
The whole algorithm is summarized in Algorithm \ref{algo_OCES} and termed as OC-ES, which stands for Optimal Control (at low-level) with Evolutionary Strategy (at high-level).

\section{Evaluations}\label{s:ch6_evaluation}

\subsubsection*{\taskref{fast_reach} - consecutive fast reaching}\manuallabel{task:fast_reach}{1}
To evaluate our proposed method, a consecutive fast reaching task is designed to test on the \MACCEPAVD\ robot.
The task requires the joint actuated by \MACCEPAVD\ to reach a sequence of three targets $\{q^*_i\}_{i=1}^3 := \{0.7,-0.35,0.3 \}$ (radians) rapidly within a fixed time horizon $T_i=1$, for $i=1,2,3$, from initial state $\bx_0 = ( 0,0,0,\pi/24,0,0 )^\T$.
The cost function $J_i$ for each subtask is defined by \eqref{eq:ch6_fastreach_example_J} - \eqref{eq:ch6_fastreach_example_l}.
A single fast reaching problem was used by \cite{Radulescu2012} to investigate the role of variable damping for \VIAs\ when an appropriate amount of damping is needed to suppress oscillation of movements. 

For comparison, a benchmark is generated by using \ILQR\ to solve the sub-problems sequentially.
The spring preset $p_s^{(i)}=\pi/24\,\mathrm{rad}$ is the lower bound of stiffness motor position command and weighting parameter $w_e^{(i)}=1$ for $i=1,2,3$. The resulting (approximately) optimal trajectory $\mathcal{S}$ is denoted by ILQR-0, and used as initial trajectory later for our proposed method. 


\subsection{Task 1: policy improvement with parametrized trajectory}
The competing terms in the composite cost function may hinder the fulfilment of all sub-goals. To investigate this issue we directly optimize the trajectory and stiffness profile simultaneously w.r.t. the composite cost function of \textbf{Task 1}
\begin{align}
J(\mathcal{S}) &= \sum_{i=1}^{3} J_i ( \mathcal{M}_i) \\
J_i &= 1000\, ((q(t_f) - q^*_i )^2 +  \dot{q}^2(t_f) )  \\
& \quad + \int_{t_0}^{t_f}  1000\,(q(t)-q^*)^2 dt \label{eq:ch6_dmpes_Jtask}  \\
& \quad  + \int_{t_0}^{t_f} (100\,(\theta_1 - q^*_i)^2 + 100 ~\theta_2^2 + 10^{-3}\,\theta_3 )\dt 
\end{align}

\begin{figure*}[!htb]
	\centering
	\begin{overpic}[]{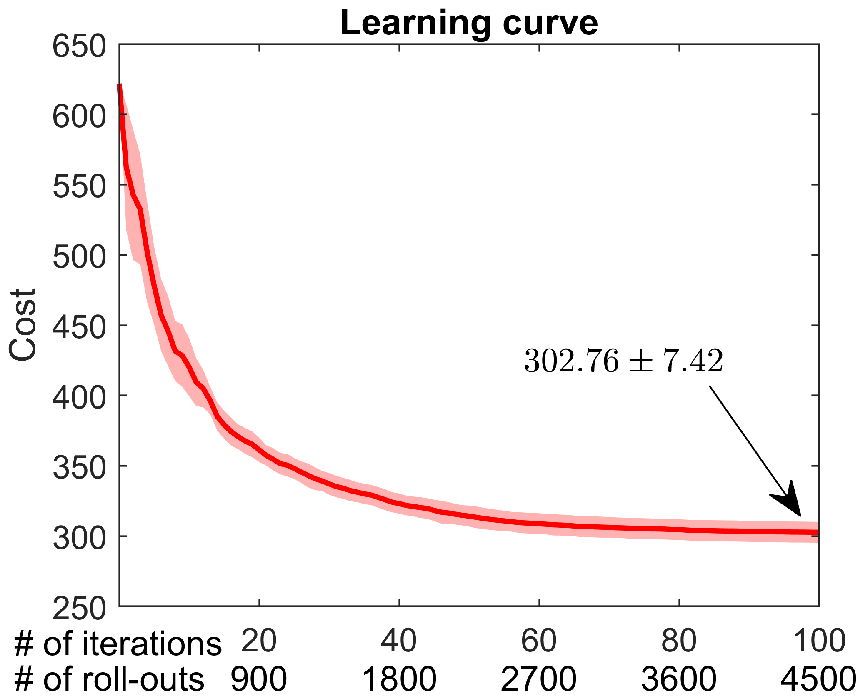}
	\end{overpic}
	\begin{overpic}[]{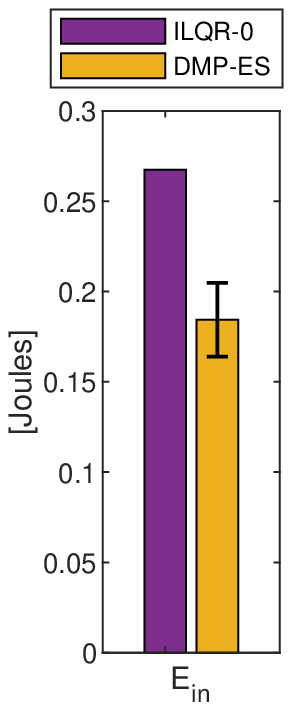}
		
	\end{overpic}
	\caption[]{\label{f:pi2seq_result}Learning curve (\textbf{left}) of $\mathrm{PI}^2\mathrm{SEQ}$ for the consecutive fast reaching task. The solid red curve is the mean of 10 runs with shaded area indicating the standard deviation. Comparison of the final energy cost with the ILQR-0 trajectory is plotted in the bar chart (\textbf{Right}). The estimated input energy cost is $0.1843 \pm 0.0204\,\mathrm{J}$ compared to ILQR-0's $0.2674\,\mathrm{J}$.}
\end{figure*}

\begin{figure}[!htb]
	\centering
	\includegraphics[]{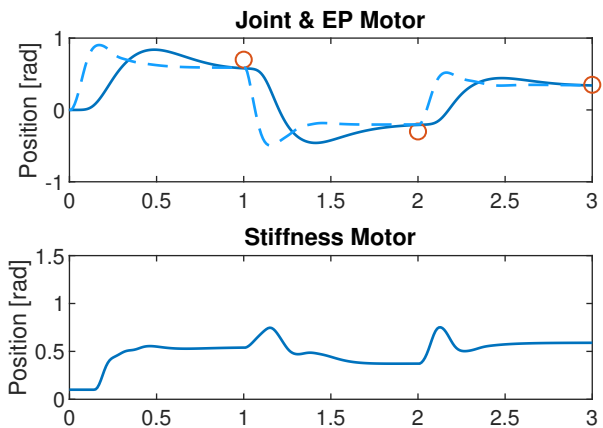}
	\caption[]{\label{f:pi2seq_traj}The final trajectory (one of 10 runs) of $\mathrm{PI}^2\mathrm{SEQ}$ for the consecutive fast reaching task. (\textbf{Top}) Joint (solid) and EP motor (dashed) trajectories, while red dots denote the targets. (\textbf{Bottom}) Stiffness motor trajectory.}
\end{figure}

The trajectories are parametrized by \DMPs\ as introduced in Appendix B. 
Each sub-movement consists of 3 \DMPs\ representing the trajectories of EP motor, stiffness motor and damping command.
All DMP are initialized with shaping parameter $\bw=\bO$, which is a $10$ dimensional vector. The goals $\bg_1,\bg_2,\bg_3$, for EP, stiffness motor, and damping respectively, are initialized as $\bg_1=\bq^*$, $\bg_2=24/\pi ~\be^{(3)}$, $\bg_3 = 0.5 ~ \be^{(3)}$,\endnote{$\bg_2$ here is actually $\bp_s$.} where $\be^{(d)}$ represents a $d$-dimensional unit vector.
The shaping parameter $\bw$ is unconstrained. 
The box constraints are  $[-\pi/3,\pi/3], [\pi/24,\pi/2],[0,1]$ for elements of $\bg_1$, $\bg_2$ and $\bg_3$ respectively.
The overall policy parameter $\bxi$ is a $99$-dimensional stacked vector of $\bg$ and $\bw$ of all three sub-movements.
Relevant meta-parameters of the algorithm are $\gamma=0.95,\mu=15,K=45$ and $\Sigma_{\bepsilon} =\mathrm{diag} (10 ~ \be^{(90)}, 0.5~\be^{(9)} ) \in \mathbb{R}^{99 \times 99}$, 
Both goals and shaping parameters $\bw$ of \DMPs\ are optimized simultaneously by Algorithm \ref{algo_OCES} except that it doesn't have an inner loop.
The policy update rule used is the same as the weighted averaging method \eqref{eq:ch6_update_normal}-\eqref{eq:ch6_update_update}. Different from what suggested by \cite{Stulp2012}, where the policy update takes the cost-to-go of sub-movements, we use the episodic cost along whole trajectory for policy update.

The learning results of $10$ sessions are illustrated in \fref{pi2seq_result}. The policy update gradually converges and final energy cost evaluated by input mechanical work $\Ein$ is successfully reduced from $0.2674\,\mathrm{J}$ of ILQR-0 to $0.1843\pm 0.0204\,\mathrm{J}$. The final trajectory of one learning session shown in \fref{pi2seq_traj} demonstrates that an optimized $\bp_s$ regulates the pretension at the transition phases. The effectiveness of policy improvement with parametrized trajectories for exploiting variable impedance of \VIAs\ is verified despite some drawbacks. 
First it can be seen that the learning takes thousands of (trajectory) samples due to high dimensionality of $\bxi$, which makes it less likely to be executed on the physical robot in an online fashion. 
Secondly, the joint trajectory in \fref{pi2seq_traj} slightly but visibly deviated from the first two goals, because the integral term in \eqref{eq:ch6_dmpes_Jtask} competes with the terminal cost of its previous movement. To circumvent this issue the composite cost function needs redesign to adjust the cost terms, weights, or impose extra constraints.

\subsection{Task 1: sequential reaching with OC-ES}
Now we take both weighting and stiffness parameters $\bw_e,\bp_s$ into account and employ the OC-ES framework.
The policy parameter $\bxi$ for \textbf{Task 1} consists of weights of control effort term and stiffness motor preset of each sub-problem. $\bxi$ is initialized as 
$\{	w_e^{(i)}=1, ~ p_s^{(i)} = \pi/24\,\mathrm{rad} \}_{i=1}^3$.
$K=4$ roll-outs are run for each policy update up to 100 iterations. 
The exploration noise $\Sigma_{\bepsilon}=0.5 ~ \bI \in \mathbb{R}^{6 \times 6}$ and decay factor $\gamma$ is set to $0.95$. 
The sample reuse parameter is chosen to be $\mu=3$.
The initial trajectory is evaluated to record its energy cost $\Ein^{(0)}$ and $J_p^{(0)}$. The latter decides the upper bound constraint of reaching performance $\bar{J}_p$ with tolerance factor $\sigma_\mathrm{tol}=0.1$.




During each roll-out, $ w_e^{(i)[k]} $ is used to set the weight of control effort term in \eqref{eq:ch6_fastreach_example_l} for $i$-th OCP, and $ p_s^{(i)[k]} $ specifies the minimal position command $u^{\mathrm{min}}_2$ of the stiffness motor. By doing so, it constrains the minimal pretension upon reaching the target. 
The sub-problems are solved by \ILQR.
After each policy update, the movement without perturbation is evaluated to record the learning performance. To verify the improvement of energy saving, both initial and final trajectories are executed on the hardware to record the energy consumption. The results are summarized in \fref{ilqrseq1_result} where ILQR-ES denotes the final trajectory.



\begin{figure*}[!htb]
	\centering
	\begin{overpic}[scale=0.9]{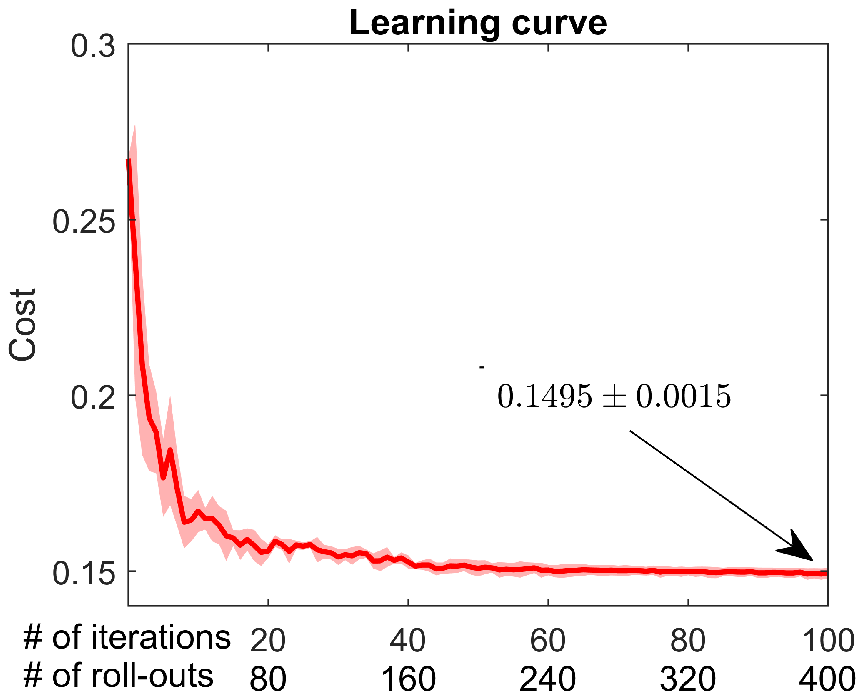}
		\put(50,-8){\ref{f:ilqrseq1_lcurve}}
	\end{overpic}
	\begin{overpic}[scale=0.9]{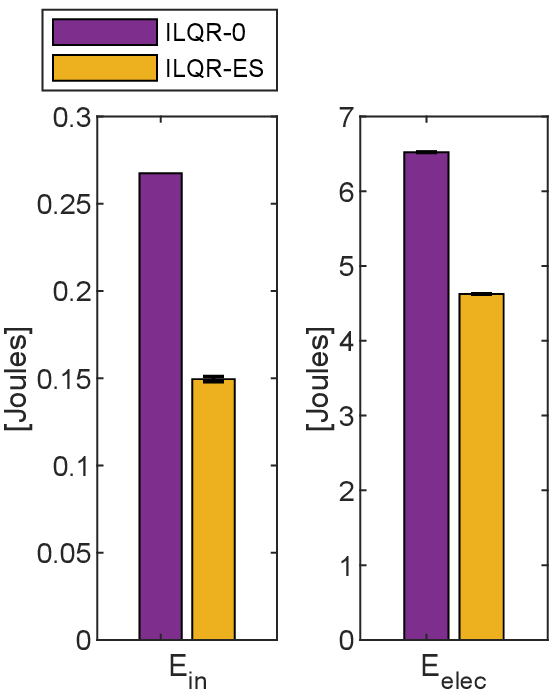}
		\put(23,-8){\ref{f:ilqrseq1_Ein}} \put(16,75){\scriptsize Simulation}
		\put(60,-8){\ref{f:ilqrseq1_Eelec}} \put(56,76){\scriptsize Record}
	\end{overpic}
	\begin{overpic}[scale=0.9]{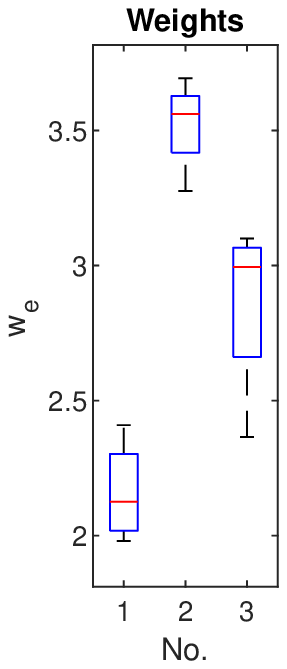}
		\put(24,-8){\ref{f:ilqrseq1_weights}}
	\end{overpic}
	\vspace{5mm}
	\caption[]{\label{f:ilqrseq1_result} Shown are:
		\begin{enumerate*}[label=(\alph*)]
			\item \label{f:ilqrseq1_lcurve} Learning curve of ILQR-ES for the consecutive fast reaching task. The solid red curve is the mean of 4 runs with shaded area indicating the standard deviation.
			\item \label{f:ilqrseq1_Ein} Estimated input energy cost of final result is $0.1495\pm 0.0015\,\mathrm{J}$ compared to $0.2674\,\mathrm{J}$ of ILQR-0.
			\item \label{f:ilqrseq1_Eelec} Electrical consumption measured on servomotors is $6.5211 \pm 0.2593 \,\mathrm{J}$, 
			while the benchmark ILQR-0 consumes $4.6261 \pm 0.2812 \,\mathrm{J}$ , which means a $29.6\%$ reduction.
			\item \label{f:ilqrseq1_weights} Distribution of optimal $w_e$ for each sub-movement.
		\end{enumerate*}  
	}
\end{figure*}

\subsubsection{Significant energy reduction}
The learning curve in \fref{ilqrseq1_result}\ref{f:ilqrseq1_lcurve} shows a fast convergence after $50$ iterations and very small variations as also shown in \fref{ilqrseq1_result}\ref{f:ilqrseq1_Ein}\ref{f:ilqrseq1_Eelec}.
It demonstrates that the OC-ES method successfully reduced the energy cost of the whole task whilst keeping worst performance cost within tolerance.
The mechanical energy cost in simulation is decreased about $44\%$ from that of the initial ILQR-0 trajectory. The electrical consumption recorded on the servomotors verifies the result with a $29.6\%$ reduction.

\subsubsection{Optimally tuned cost function weighting}
Looking at \fref{ilqrseq1_result}\ref{f:ilqrseq1_weights}, all $\bw_e$ of sub-problems tend to increase from the initial settings. Despite relative large variations of the optimization result, it can be observed that the second sub-movement takes the highest weight for control cost, which indicates the energy efficiency of it is most critical. In \fref{ilqrseq1_traj} we can see that the second sub-movement has the largest travel distance among the three, and consumes the most energy in the initial trajectory (as shown by the accumulated energy cost in the right column in \fref{ilqrseq1_traj}). Hence, the result can be explained as the optimization adjusts the weight to balance the performance-cost trade-off more towards reducing energy cost.

\subsubsection{Exploiting variable stiffness}
It can be seen in \fref{ilqrseq1_traj} that energy reduction occurs significantly during the second and third movements, compared with the initial trajectory. 
The stiffness motor maintains higher pretension at transition phases (\fref{ilqrseq1_traj}(f)) due to the constraint imposed by optimized $\bp_s$, by which the acceleration of the subsequent movement consumes less energy in the EP motor. 
Also, the adjustment of stiffness motor causes a lot of electrical consumption (\fref{ilqrseq1_traj}(h)), suggesting that the control effort may be lead to suboptimal solutions regarding real energy consumption. However, this highly depends on the variable stiffness mechanism and hardware design. For example, by implementing the \VSAs\ designed for minimizing energy cost for stiffness modulation (\cite{Jafari2015,Chalvet2017}), the energy cost of the stiffness motor of initial trajectory can be reduced so that the most saving occurs on the EP motor. However, it would raise another problem that if a \VSA\ does not require energy input to adjust stiffness, then it may not be able to pre-store energy at equilibrium position (EP). As a result there may be no energy buffering effect for some movements starting from a static equilibrium phase.



\begin{figure*}[!htb]
	\centering
	\begin{overpic}[width=1\linewidth]{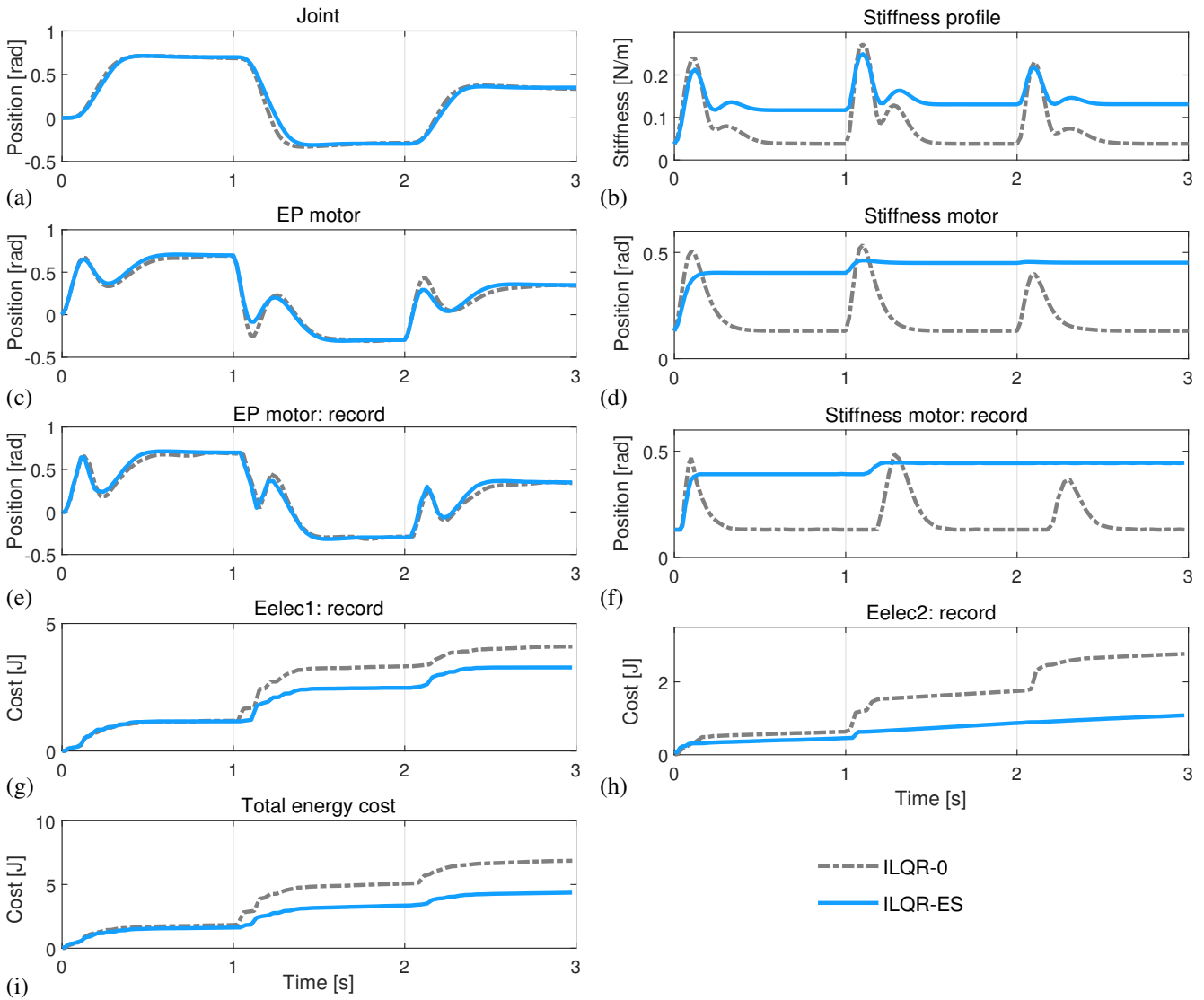}
		\put(0,67){(a)}\put(50,67){(b)}
		\put(0,50){(c)}\put(50,50){(d)}
		\put(0,33){(e)}\put(50,33){(f)}
		\put(0,17){(g)}\put(50,17){(h)}
		\put(0,0){(i)}
	\end{overpic}
	\caption[]{\label{f:ilqrseq1_traj}Result of executing the trajectory in simulation and on real hardware of ILQR-ES for consecutive fast reaching, compared with ILQR-0 trajectory as a benchmark. ILQR-0 also serves as the initial trajectory. The last two rows show the measured electrical cost by cumulating the recorded power along the trajectory.}
\end{figure*}

Overall, the experiment demonstrates the effectiveness of applying OC-ES framework to improve energy efficiency by exploiting variable stiffness and cost function tuning. The learning takes only $4$ explorations per iteration by leveraging model-based \OC\ at the low-level, making it more feasible to run on the real robot.

\subsection{Temporal and Stiffness Optimization for Tracking Control}
The second application is to show that the proposed framework can be applied to temporal optimization and work with a low-level tracking controller.

\subsubsection*{\taskref{tracking} - Consecutive trajectory tracking}\manuallabel{task:tracking}{2}
This task requires the arm to smoothly reach a sequence of targets with minimal-jerk joint trajectory. In addition to exploiting variable stiffness, the relative timing is allowed to be optimized. We set the targets as $\{q^*_i \} = [\pi/5, -0.2, 1,0.3]$ ($\mathrm{rad}$). The arm starts at $q(0)=0 \,\mathrm{rad}$. The total time for the movement is $2.4 \, \mathrm{s}$.
$\bxi$ is defined as $\bxi = ( \bt_d^\T, \bp_s^\T )^\T \in \mathbb{R}^7$, where $\bt_d=\{t^{(i)}_d\}_{i=1}^{3}, \bp_s=\{p_s^{(i)}\}_{i=1}^{4}$.
Since the total time is kept the same, the last time duration is excluded from the policy parameter. 
The box constraint on $\bxi$ is
\begin{align}
\bxi_\mathrm{min} = (0.3,0.3,0.3,0,0,0,0)^\T, \nonumber \\ \bxi_\mathrm{max}=(1.2,1.2,1.2,\frac{\pi}{2},\frac{\pi}{2},\frac{\pi}{2},\frac{\pi}{2})^\T \nonumber
\end{align}

\begin{figure*}[!htb]
	\centering
	\begin{overpic}[]{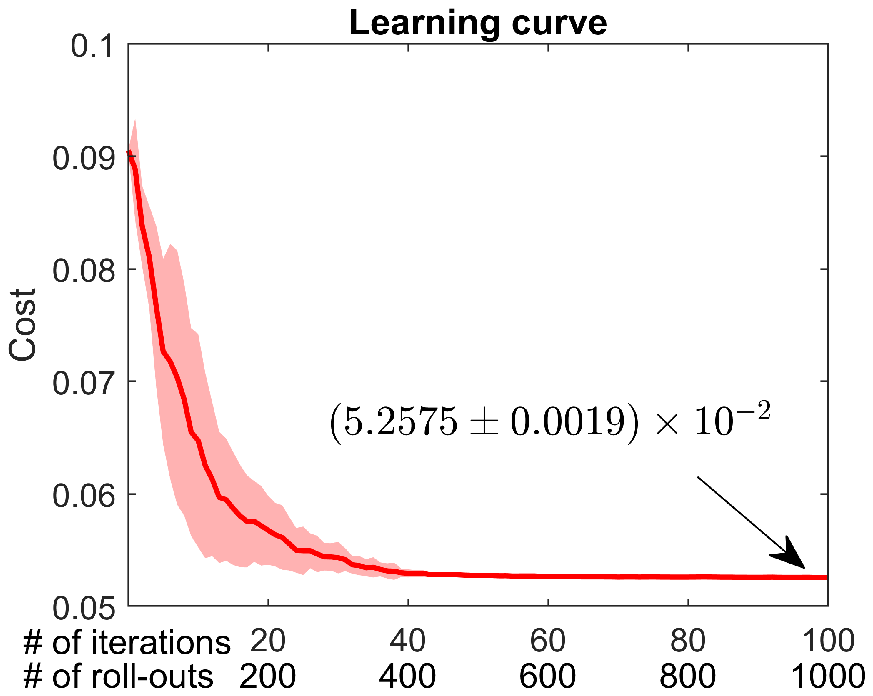}
	\end{overpic}
	\begin{overpic}[]{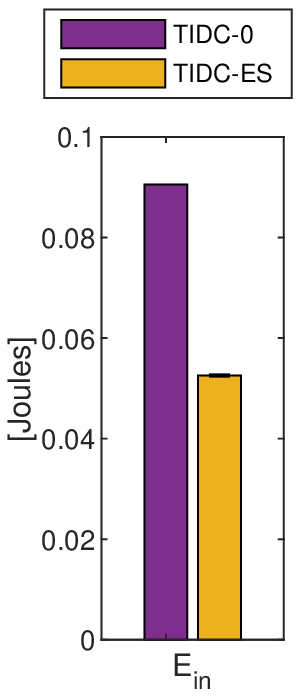}
		
	\end{overpic}
	\caption[]{\label{f:ch6_eidcseq_result}Learning curve (\textbf{left}) of TIDC-ES for the consecutive fast reaching task. The solid red curve is the mean of 10 runs with shaded area indicating the standard deviation. Comparison of the final energy cost with the TIDC-0 trajectory is plotted in the bar chart (\textbf{right}). The estimated input energy cost is $(5.2575 \pm 0.0019) \times 10^{-2}\,\mathrm{J}$ compared to TIDC-0's $9.054 \times 10^{-2} \,\mathrm{J}$.}
\end{figure*}

The minimal-jerk joint trajectory can be computed analytically by formula introduced in \cite{Flash1985}, given the time duration $\bt_d$ and where it begins and ends. Then it becomes a joint space tracking problem. The joint tracking with extended inverse dynamics controller (TIDC) derived in Appendix C to track the joint trajectory and resolve the actuation redundancy automatically.
The controller serves as a feedback control law and reduces the inner loop OCP to a forward pass of dynamics. 
The stiffness parameter $\bp_s$ is used to impose a constraint on the target position of stiffness motor in each sub-movement, by adding a null-space controller 
\begin{equation}
\bv_{\mathrm{ns}}= ( (q^*_i - \theta_1) , p_s^{(i)} - \theta_2, 0 )^\T
\end{equation}
for $i$-th trajectory tracking. This null-space controller encourages the EP motor to move towards the joint target and the stiffness motor to $p_s^{(i)}$.
Other relevant meta-parameters for the policy improvement method are: $K=10,\mu=3,\gamma=0.97,\sigma_{\mathrm{tol}}=0.01,\Sigma_{\bepsilon}=\mathrm{diag}  ( 0.3\,\be^{(3)}  , 0.5 \,\be^{(4)} )$. The whole method is termed as TIDC-ES.

\begin{table*}[!htb]
	\centering
	\begin{tabular}{ c||c|c|c|c }
		\hline 
		\multicolumn{5}{c}{Optimized parameters} \\
		\hline
		No. & 1 & 2 & 3 & 4 \\
		\hline 
		$t^{(i)}_d \, [\mathrm{ms}]$ & $555.8\pm 5.1$ & $593.1 \pm 2.5$ & $692.6\pm 3.4 $& $558.5 \pm1.6 $ \\
		\hline
		$p_s^{(i)} [\mathrm{rad}]$ & $1.259 \pm 0.013$ & $0.782\pm 0.007 $ & $0.412\pm 0.008$ & $ 0.002 \pm 0.002 $ \\
		\hline	
	\end{tabular}
	\caption[Optimized parameters of temporal and stiffness optimization with TIDC-ES.]{\label{t:ch6_eidcseq_timestiff} Optimized parameters of temporal and stiffness optimization with TIDC-ES. }
\end{table*}

The initial trajectory is generated with
\[ \bxi^{(0)}=(0.6,0.6,0.6,0.2,0.2,0.2,0.2)^\T \]
and denoted by TIDC-0. The learning results after 100 iterations are presented in \fref{ch6_eidcseq_result}. It can be seen that the learning curve initially has a large variation but quickly converges after 40 iterations. Compared to the initial trajectory, by exploiting stiffness and temporal optimization the input mechanical energy $\Ein$ reduces by about $42\%$. 
Looking at the results in \tref{ch6_eidcseq_timestiff}, 
the optimized stiffness targets range from $1.26\,\mathrm{rad}$ for the first sub-movement to nearly $0 \,\mathrm{rad}$ for the last one. It results in the pretension increasing during the first two sub-movements then decreasing towards the end (as shown in \fref{ch6_eidcseq_stiffness}).
Moreover, the duration of third sub-movement is optimized to $692.6\,\mathrm{ms}$, which is $92.6\,\mathrm{ms}$ more than the initial setting. While other three sub-movements have shorter durations. The result is coherent with the order in terms of movement distance.

\begin{figure}[!htb]
	\centering
	\includegraphics[]{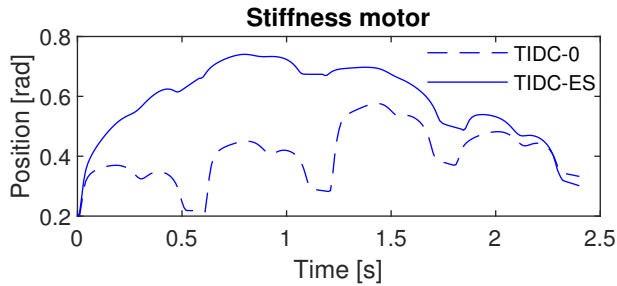}
	\caption[]{\label{f:ch6_eidcseq_stiffness} Stiffness motor profile of TIDC-ES final trajectory (blue solid) compared to initial trajectory (dashed blue).}
\end{figure}

\subsection{Discussion}
The experiments presented in this section demonstrated noticeable energy saving realized in consecutive reaching tasks.
The task was the same as the one used in \cite{Wu2020},
although there a default spring pretension was chosen manually for all movements.

The proposed method in this paper has been demonstrated to help the robot automatically regulate its stiffness with awareness of subsequent movements. 
The result of ILQR-ES for \taskref{fast_reach} regulates the stiffness motor to maintain at a small range around $0.5\,\mathrm{rad}$, suggesting that a fixed value can be tuned for energy efficiency in practice if the movement distances are not distributed diversely. 
In general, it suggests that for \VIAs\ that rely on spring pretension to modulate stiffness, the more efficient way to use them in consecutive point-to-point reaching is not to reset their stiffness to minimum by default. 
Hence, using control effort to represent energy cost is questionable as it encourages the stiffness to move towards the minimum at the end of each submovement.
However, due to the fact that accurate estimation of $E_{\mathrm{in}},E_\mathrm{elec}$ needs a much smaller time step for discretization of the continuous system dynamics, the quadratic control effort is preferred for less computation cost. It also enhances smoothness of the trajectory, although at a cost to energetic optimality. Nevertheless, this loss is alleviated --- by applying the proposed framework --- with an upper layer optimizer that adjusts the trade-off balance.




\section{Conclusions}\label{s:ch6_conclusion}\noindent
This paper proposed a versatile framework that integrates Optimal Control and Evolution Strategy (OC-ES) in a bi-level structure to address the optimization of movement sequence specifically for \VIAs. At the low-level OC is leveraged to resolve the actuation redundancy and exploit variable impedance naturally.
The high-level optimization in sequential context is formulated as a reinforcement learning problem, in the form of iterative policy improvement, and solved as a black box optimization using method inspired by evolutionary strategy.

The proposed framework was applied for two consecutive reaching tasks on a \MACCEPAVD\ actuator, one requires reaching as quickly as possible, the other tracks a smooth trajectory in joint space.
In both cases natural dynamics is hard to be exploited for energy buffering as in periodic movement.
By investigating the performance-cost trade-off via efficient frontiers, it can be seen how cost function weighting and minimal stiffness preset influence the energy efficiency.
These two aspects can be addressed in the sequential movement context via the proposed framework, by which variable impedance can be fully exploited and the low-level trade-off is optimally balanced. 
In addition, a tracking controller that resolves the actuation redundancy was implemented to show the temporal and stiffness optimization at high-level can improve the energy efficiency of low-level sequential tracking control.
All the experiments presented in \sref{ch6_evaluation} demonstrated significant improvement of energy efficiency in both simulations and on hardware.

However, this work has been limited to reaching movements. More task types need to be considered in the future work to demonstrate more complex behaviours. Furthermore, it would be interesting to extend the application to compliant robots with multiple DOFs and consider problems involving contacts and interactions.

\theendnotes

\bibliographystyle{SageH}
\bibliography{mybib2020.bib}

\begin{thebibliography}{52}
\providecommand{\natexlab}[1]{#1}
\providecommand{\url}[1]{\texttt{#1}}
\providecommand{\urlprefix}{URL }
\expandafter\ifx\csname urlstyle\endcsname\relax
  \providecommand{\doi}[1]{DOI:\discretionary{}{}{}#1}\else
  \providecommand{\doi}{DOI:\discretionary{}{}{}\begingroup
  \urlstyle{rm}\Url}\fi

\bibitem[{Ariani and Diedrichsen(2019)}]{Ariani2019}
Ariani G and Diedrichsen J (2019) {Sequence learning is driven by improvements
  in motor planning}.
\newblock \emph{Journal of Neurophysiology} 121(6): 2088--2100.
\newblock \doi{10.1152/jn.00041.2019}.

\bibitem[{Berret et~al.(2011)Berret, Ivaldi, Nori and Sandini}]{Berret2011}
Berret B, Ivaldi S, Nori F and Sandini G (2011) {Stochastic optimal control
  with variable impedance manipulators in presence of uncertainties and delayed
  feedback}.
\newblock \emph{IEEE International Conference on Intelligent Robots and
  Systems} : 4354--4359\doi{10.1109/IROS.2011.6048586}.

\bibitem[{Bobbert(2001)}]{Bobbert2001}
Bobbert MF (2001) {Dependence of human squat jump performance on the series
  elastic compliance of the triceps surae: A simulation study}.
\newblock \emph{Journal of Experimental Biology} 204(3): 533--542.

\bibitem[{Braun et~al.(2012)Braun, Howard and Vijayakumar}]{Braun2012}
Braun D, Howard M and Vijayakumar S (2012) Optimal variable stiffness control:
  formulation and application to explosive movement tasks.
\newblock \emph{Autonomous Robots} 33(3): 237--253.

\bibitem[{Braun et~al.(2013)Braun, Petit, Huber, Haddadin, Van Der~Smagt,
  Albu-Schaffer and Vijayakumar}]{Braun2013}
Braun D, Petit F, Huber F, Haddadin S, Van Der~Smagt P, Albu-Schaffer A and
  Vijayakumar S (2013) {Robots driven by compliant actuators: Optimal control
  under actuation constraints}.
\newblock \emph{IEEE Transactions on Robotics} 29(5): 1085--1101.

\bibitem[{Burridge et~al.(1999)Burridge, Rizzi and Koditschek}]{Burridge1999}
Burridge RR, Rizzi AA and Koditschek DE (1999) {Sequential composition of
  dynamically dexterous robot behaviors}.
\newblock \emph{International Journal of Robotics Research} 18(6): 534--555.
\newblock \doi{10.1177/02783649922066385}.

\bibitem[{Chalvet and Braun(2017)}]{Chalvet2017}
Chalvet V and Braun DJ (2017) {Criterion for the Design of Low-Power Variable
  Stiffness Mechanisms}.
\newblock \emph{IEEE Transactions on Robotics} 33(4): 1002--1010.
\newblock \doi{10.1109/TRO.2017.2689068}.

\bibitem[{Conn et~al.(2009)Conn, Scheinberg and Vicente}]{Conn2009}
Conn AR, Scheinberg K and Vicente LN (2009) \emph{{Introduction to
  Derivative-Free Optimization}}.
\newblock SIAM.

\bibitem[{Flash and Hogan(1985)}]{Flash1985}
Flash T and Hogan N (1985) {The coordination of arm movements: an
  experimentally confirmed mathematical model}.
\newblock \emph{The Journal of neuroscience} 5(7): 1688--703.
\newblock \doi{4020415}.

\bibitem[{Haddadin et~al.(2018)Haddadin, Krieger, Albu-Schaffer and
  Lilge}]{Haddadin2018}
Haddadin S, Krieger K, Albu-Schaffer A and Lilge T (2018) Exploiting elastic
  energy storage for blind cyclic manipulation: Modeling, stability analysis,
  control, and experiments for dribbling.
\newblock \emph{IEEE Transactions on Robotics} 34(1): 91--112.

\bibitem[{Hansen and Ostermeier(2001)}]{Hansen2001}
Hansen N and Ostermeier A (2001) {Completely Derandomized Self-Adaptation in
  Evolution Strategies}.
\newblock \emph{Evolutionary Computation} 9(2): 159--195.
\newblock \doi{10.1162/106365601750190398}.

\bibitem[{Hogan and Sternad(2012)}]{Hogan2012}
Hogan N and Sternad D (2012) {Dynamic primitives of motor behavior}.
\newblock \emph{Biological Cybernetics} 106(11-12): 727--739.
\newblock \doi{10.1007/s00422-012-0527-1}.

\bibitem[{Huang et~al.(2012)Huang, Kram and Ahmed}]{Huang2012}
Huang HJ, Kram R and Ahmed AA (2012) {Reduction of Metabolic Cost during Motor
  Learning of Arm Reaching Dynamics}.
\newblock \emph{The Journal of Neuroscience} 32(6): 2182--2190.
\newblock \doi{10.1523/JNEUROSCI.4003-11.2012}.

\bibitem[{Ijspeert et~al.(2002)Ijspeert, Nakanishi and Schaal}]{Ijspeert2002}
Ijspeert A, Nakanishi J and Schaal S (2002) {Movement imitation with nonlinear
  dynamical systems in humanoid robots}.
\newblock \emph{IEEE International Conference on Robotics and Automation}
  (May): 1398--1403.

\bibitem[{Ijspeert et~al.(2013)Ijspeert, Nakanishi, Hoffmann, Pastor and
  Schaal}]{Ijspeert2013}
Ijspeert AJ, Nakanishi J, Hoffmann H, Pastor P and Schaal S (2013) {Dynamical
  movement primitives: learning attractor models for motor behaviors.}
\newblock \emph{Neural computation} 25(2): 328--73.

\bibitem[{Jafari et~al.(2015)Jafari, Tsagarakis and Caldwell}]{Jafari2015}
Jafari A, Tsagarakis N and Caldwell D (2015) Energy efficient actuators with
  adjustable stiffness: a review on awas, awas-ii and compact vsa changing
  stiffness based on lever mechanism.
\newblock \emph{Industrial Robot: An International Journal} 42(3): 242--251.
\newblock \doi{10.1108/IR-12-2014-0433}.

\bibitem[{Lashley(1951)}]{Lashley1951}
Lashley KS (1951) {The problem of serial order in behavior.}
\newblock In: \emph{Cerebral mechanisms in behavior; the Hixon Symposium.}
  Oxford, England: Wiley, pp. 112--146.

\bibitem[{Lay et~al.(2002)Lay, Sparrow, Hughes and O'Dwyer}]{Lay2002}
Lay B, Sparrow W, Hughes K and O'Dwyer N (2002) {Practice effects on
  coordination and control , metabolic energy expenditure , and muscle
  activation}.
\newblock \emph{Human Movement Science} 21: 807--830.
\newblock \doi{10.1016/S0167-9457(02)00166-5}.

\bibitem[{Levine and Koltun(2012)}]{Levine2012}
Levine S and Koltun V (2012) {Continuous inverse optimal control with locally
  optimal examples}.
\newblock \emph{Proceedings of the 29th International Conference on Machine
  Learning, ICML 2012} 1: 41--48.

\bibitem[{Li and Todorov(2004)}]{Li2004}
Li W and Todorov E (2004) {Iterative Linear Quadratic Regulator Design for
  Nonlinear Biological Movement Systems}.
\newblock In: \emph{IEEE Int. Conf. Robotics \& Automation}.

\bibitem[{Lucia et~al.(2013)Lucia, Umezawa, Nakamura and Billard}]{Lucia2013}
Lucia P, Umezawa K, Nakamura Y and Billard A (2013) {Learning Robot Skills
  Through Motion Segmentation and Constraints Extraction}.
\newblock In: \emph{HRI Workshop on Collaborative Manipulation}.

\bibitem[{Manschitz et~al.(2015)Manschitz, Kober, Gienger and
  Peters}]{Manschitz2015}
Manschitz S, Kober J, Gienger M and Peters J (2015) {Learning movement
  primitive attractor goals and sequential skills from kinesthetic
  demonstrations}.
\newblock \emph{Robotics and Autonomous Systems} 74: 97--107.
\newblock \doi{10.1016/j.robot.2015.07.005}.

\bibitem[{Matsusaka et~al.(2016)Matsusaka, Uemura and Kawamura}]{Matsusaka2016}
Matsusaka K, Uemura M and Kawamura S (2016) Realization of highly energy
  efficient pick-and-place tasks using resonance-based robot motion control.
\newblock \emph{Advanced Robotics} 30(9): 608--620.

\bibitem[{Mombaur et~al.(2010)Mombaur, Truong and Laumond}]{Mombaur2010}
Mombaur K, Truong A and Laumond JP (2010) {From human to humanoid
  locomotion—an inverse optimal control approach}.
\newblock \emph{Autonomous Robots} 28(3): 369--383.
\newblock \doi{10.1007/s10514-009-9170-7}.

\bibitem[{Motegi and Matsui(2011)}]{Motegi2011}
Motegi M and Matsui T (2011) {Optimal Control Model for Reproducing Squat
  Movements Based on Successive-Movement Combination}.
\newblock \emph{The Proceedings of the Symposium on sports and human dynamics}
  2011: 558--563.
\newblock \doi{10.1299/jsmeshd.2011.558}.

\bibitem[{Nakanishi et~al.(2016)Nakanishi, Radulescu, Braun and
  Vijayakumar}]{Nakanishi2016}
Nakanishi J, Radulescu A, Braun DJ and Vijayakumar S (2016) {Spatio-temporal
  stiffness optimization with switching dynamics}.
\newblock \emph{Autonomous Robots} : 1--19.

\bibitem[{Nakanishi et~al.(2011)Nakanishi, Rawlik and
  Vijayakumar}]{Nakanishi2011}
Nakanishi J, Rawlik K and Vijayakumar S (2011) {Stiffness and temporal
  optimization in periodic movements: An optimal control approach}.
\newblock In: \emph{IEEE/RSJ International Conference on Intelligent Robots and
  Systems}, 1. pp. 718--724.

\bibitem[{Nelson(1983)}]{Nelson1983}
Nelson WL (1983) {Physical principles for economies of skilled movements.}
\newblock \emph{Biological cybernetics} 46: 135--147.
\newblock \doi{10.1007/BF00339982}.

\bibitem[{Okada et~al.(2002)Okada, Ban and Nakamura}]{Okada2002}
Okada M, Ban S and Nakamura Y (2002) {Skill of compliance with controlled
  charging/discharging of kinetic energy}.
\newblock In: \emph{Proceedings 2002 IEEE International Conference on Robotics
  and Automation (Cat. No.02CH37292)}, volume~3. IEEE.
\newblock ISBN 0-7803-7272-7, pp. 2455--2460.
\newblock \doi{10.1109/ROBOT.2002.1013600}.

\bibitem[{Radulescu et~al.(2012)Radulescu, Howard, Braun and
  Vijayakumar}]{Radulescu2012}
Radulescu A, Howard M, Braun DJ and Vijayakumar S (2012) {Exploiting variable
  physical damping in rapid movement tasks}.
\newblock In: \emph{IEEE/ASME Int. Conf. Advanced Intelligent Mechatronics}.

\bibitem[{Rawlik et~al.(2010)Rawlik, Toussaint and Vijayakumar}]{Rawlik2010}
Rawlik K, Toussaint M and Vijayakumar S (2010) {An Approximate Inference
  Approach to Temporal Optimization in Optimal Control}.
\newblock \emph{Neural Information Processing Systems} : 1--9.

\bibitem[{{Reher} et~al.(2016){Reher}, {Cousineau}, {Hereid}, {Hubicki} and
  {Ames}}]{Reher2016}
{Reher} J, {Cousineau} EA, {Hereid} A, {Hubicki} CM and {Ames} AD (2016)
  Realizing dynamic and efficient bipedal locomotion on the humanoid robot
  durus.
\newblock In: \emph{2016 IEEE International Conference on Robotics and
  Automation (ICRA)}. pp. 1794--1801.
\newblock \doi{10.1109/ICRA.2016.7487325}.

\bibitem[{Roberts(2016)}]{Roberts2016}
Roberts TJ (2016) {Contribution of elastic tissues to the mechanics and
  energetics of muscle function during movement}.
\newblock \emph{The Journal of Experimental Biology} 219(2): 266--275.
\newblock \doi{10.1242/jeb.124446}.

\bibitem[{Roozing et~al.(2016)Roozing, Li, Caldwell and
  Tsagarakis}]{Roozing2016DesignEfficiency}
Roozing W, Li Z, Caldwell DG and Tsagarakis NG (2016) {Design Optimisation and
  Control of Compliant Actuation Arrangements in Articulated Robots for
  Improved Energy Efficiency}.
\newblock \emph{IEEE Robotics and Automation Letters} 1(2).
\newblock \doi{10.1109/LRA.2016.2521926}.

\bibitem[{Roozing et~al.(2019)Roozing, Ren and Tsagarakis}]{Roozing2019}
Roozing W, Ren Z and Tsagarakis NG (2019) {An efficient leg with
  series–parallel and biarticular compliant actuation: design optimization,
  modeling, and control of the eLeg}.
\newblock \emph{International Journal of Robotics Research}
  \doi{10.1177/0278364919893762}.

\bibitem[{Schaal(2006)}]{Schaal2006}
Schaal S (2006) {Dynamic Movement Primitives – A Framework for Motor Control
  in Humans and Humanoid Robotics}.
\newblock \emph{Adaptive Motion of Animals and Machines} (1): 261--280.
\newblock \doi{10.1007/4-431-31381-8_23}.

\bibitem[{Schaal and Atkeson(2010)}]{Schaal2010}
Schaal S and Atkeson CG (2010) {Learning control in robotics}.
\newblock \emph{IEEE Robotics and Automation Magazine} 17(2): 20--29.
\newblock \doi{10.1109/MRA.2010.936957}.

\bibitem[{Sparrow and Newell(1998)}]{Sparrow1998}
Sparrow W and Newell KM (1998) {Metabolic energy expenditure and the regulation
  ofmovement economy}.
\newblock \emph{Psychonomic Bulletin {\&} Review} 5(2): 173--196.

\bibitem[{Stulp and Sigaud(2013)}]{Stulp2013}
Stulp F and Sigaud O (2013) {Robot Skill Learning: From Reinforcement Learning
  to Evolution Strategies}.
\newblock \emph{Paladyn, Journal of Behavioral Robotics} 4(1).
\newblock \doi{10.2478/pjbr-2013-0003}.

\bibitem[{Stulp et~al.(2012)Stulp, Theodorou and Schaal}]{Stulp2012}
Stulp F, Theodorou EA and Schaal S (2012) {Reinforcement Learning With
  Sequences of Motion Primitives for Robust Manipulation}.
\newblock \emph{IEEE Transactions on Robotics} 28(6): 1360--1370.

\bibitem[{Tassa et~al.(2014)Tassa, Mansard and Todorov}]{Tassa2014}
Tassa Y, Mansard N and Todorov E (2014) {Control-limited differential dynamic
  programming}.
\newblock In: \emph{ICRA}. pp. 1168--1175.

\bibitem[{Theodorou et~al.(2010)Theodorou, Buchli and
  Schaal}]{Theodorou2010JMLR}
Theodorou E, Buchli J and Schaal S (2010) A generalized path integral control
  approach to reinforcement learning.
\newblock \emph{J. Mach. Learn. Res.} 11: 3137–3181.

\bibitem[{Todorov and Jordan(2002)}]{Todorov2002}
Todorov E and Jordan M (2002) {Optimal feedback control as a theory of motor
  coordination}.
\newblock \emph{Naturenature Neuroscience} (5): 1226--1235.
\newblock \doi{https://doi.org/10.1038/nn963}.

\bibitem[{Toussaint et~al.(2007)Toussaint, Gienger and Goerick}]{Toussaint2007}
Toussaint M, Gienger M and Goerick C (2007) {Optimization of sequential
  attractor-based movement for compact behaviour generation}.
\newblock In: \emph{2007 7th IEEE-RAS International Conference on Humanoid
  Robots}, 2. IEEE.
\newblock ISBN 978-1-4244-1861-9, pp. 122--129.
\newblock \doi{10.1109/ICHR.2007.4813858}.

\bibitem[{{Tsagarakis} et~al.(2013){Tsagarakis}, {Morfey}, {Medrano Cerda},
  {Zhibin} and {Caldwell}}]{Tsagarakis2013}
{Tsagarakis} NG, {Morfey} S, {Medrano Cerda} G, {Zhibin} L and {Caldwell} DG
  (2013) Compliant humanoid coman: Optimal joint stiffness tuning for modal
  frequency control.
\newblock In: \emph{2013 IEEE International Conference on Robotics and
  Automation}. pp. 673--678.

\bibitem[{Van~Ham et~al.(2007)Van~Ham, Vanderborght, Van~Damme, Verrelst and
  Lefeber}]{VanHam2007}
Van~Ham R, Vanderborght B, Van~Damme M, Verrelst B and Lefeber D (2007)
  {MACCEPA, the mechanically adjustable compliance and controllable equilibrium
  position actuator: Design and implementation in a biped robot}.
\newblock \emph{Rob. Auton. Syst.} 55(10): 761--768.
\newblock \doi{10.1016/j.robot.2007.03.001}.

\bibitem[{Verstraten et~al.(2016)Verstraten, Furnemont, Mathijssen,
  Vanderborght and Lefeber}]{Verstraten2016}
Verstraten T, Furnemont R, Mathijssen G, Vanderborght B and Lefeber D (2016)
  {Energy Consumption of Geared DC Motors in Dynamic Applications: Comparing
  Modeling Approaches}.
\newblock \emph{IEEE Robotics and Automation Letters} 1(1): 524--530.
\newblock \doi{10.1109/LRA.2016.2517820}.

\bibitem[{Wilson and Flanagan(2008)}]{Wilson2008}
Wilson JM and Flanagan EP (2008) {The Role of Elastic Energy in Activities with
  High Force and Power Requirements: A Brief Review}.
\newblock \emph{Journal of Strength and Conditioning Research} 22(5):
  1705--1715.
\newblock \doi{10.1519/JSC.0b013e31817ae4a7}.

\bibitem[{Wolf and Hirzinger(2008)}]{Wolf2008}
Wolf S and Hirzinger G (2008) {A new variable stiffness design: Matching
  requirements of the next robot generation}.
\newblock In: \emph{2008 IEEE International Conference on Robotics and
  Automation}. IEEE, pp. 1741--1746.

\bibitem[{{Wu} and {Howard}(2020)}]{Wu2020}
{Wu} F and {Howard} M (2020) Energy regenerative damping in variable impedance
  actuators for long-term robotic deployment.
\newblock \emph{IEEE Transactions on Robotics} : 1--13.

\bibitem[{Yokoi and Diedrichsen(2019)}]{Yokoi2019}
Yokoi A and Diedrichsen J (2019) {Neural Organization of Hierarchical Motor
  Sequence Representations in the Human Neocortex}.
\newblock \emph{Neuron} 103(6): 1178 -- 1190.e7.
\newblock \doi{https://doi.org/10.1016/j.neuron.2019.06.017}.

\bibitem[{Yuan(2015)}]{Yuan2015}
Yuan Yx (2015) {Recent advances in trust region algorithms}.
\newblock \emph{Mathematical Programming} 151(1): 249--281.
\newblock \doi{10.1007/s10107-015-0893-2}.

\end{thebibliography}


\section*{Appendix A: The robot model}\label{app:maccepavd}

The forward dynamics of MACCEPAVD can be written as:
\begin{align}
\ddot{q}    &= (\tau_s - d(u_3) \qdot - b\qdot -\tau_{\mathrm{ext}}) \inertia^{-1} \\\label{e:motor-dynamics1}
\ddot{\theta}_1 &= \beta^2(u_1 - \theta_1) - 2\beta\dot{\theta}_1    \\\label{e:motor-dynamics2}
\ddot{\theta}_2 &= \beta^2(u_2 - \theta_2) - 2\beta\dot{\theta}_2 
\end{align}
where $q, \dot{q}, \ddot{q}$ are the joint angle, velocity and acceleration, respectively, $b$ is the viscous friction coefficient for the joint, $\inertia$ is the link inertia, $\tau_s$ is the torque generated by the spring force, and $\tau_{\mathrm{ext}}$ is the joint torque due to external loading (the following reports results for the case of no external loading, \ie $\tau_{\mathrm{ext}}=0$). $\theta_1, \theta_2, \dot{\theta}_1, \dot{\theta}_2, \ddot{\theta}_1, \ddot{\theta}_2$ are the motor angles, velocities and accelerations. 

The motor angles $\theta_1,\theta_2$ and damping $d$ are controlled by control input $\bu=(u_1,u_2,u_3)^\T$.
The servomotor dynamics \eref{motor-dynamics1}, \eref{motor-dynamics2} are assumed to behave as a critically damped system,
with $\beta$ constraining the maximum acceleration of the 2nd order dynamical system.

The torque $\tau_s$ can be calculated as follows:
\begin{align}
\tau_s &= \kappa B C \sin{ (\theta_1 - q) } (1+ \frac{r \theta_2 - |C-B|}{A(q,\theta_1)}) \\
\tau_{l_1} & = \tau_s \\
\tau_{l_2} &= \kappa (r \theta_2 - |C-B| +A(q,\theta_1))
\end{align}
where $A(q,\theta_1) =\sqrt{B^2+C^2-2BC\cos{ (\theta_1 - q) }}$, $B$ and $C$ are the lengths shown in \fref{maccepavd}, $r$ is the radius of the winding drum used to adjust the spring pre-tension, and $\kappa$ is the linear spring constant. 


The damping coefficient $d(u_3)$ linearly depends on control input $u_3$ and
\begin{equation}
d(u_3) = \bar{d} u_3,
\end{equation}
where $\bar{d}$ is maximum damping coefficient and the control input varies from $0$ to $1$ ($u_3 \in [0,1]$). 

The mechanical and electrical power of motor $i$ are estimated by
\begin{align}
	P_{\mathrm{in},i} &= \tau_{l,i}  \dot{\theta}_i \\
	P_{\mathrm{elec},i} &= (\frac{\tau_{m,i}}{n_g k})^2 R_m +  [ J_m \ddot{\theta}_i \, \dot{\theta}_i ]^+ + b_f \dot{\theta}_i^2 + [\tau_{l,i} \dot{\theta}_i]^+ \\
	\tau_{m,i} &= \tau_{l,i} + J_{m} \ddot{\theta}_i + b_{f} \dot{\theta}_i 
\end{align}

On the hardware, SERVO1 and SERVO2 are two Robotis Dynamixel XM430-210-R servomotors with internal position and current sensors. The sensing data is transmitted from servos to a PC hosting connected with a dedicated U2D2 USB converter. The communication between servos and PC is based on ROS messages.

The corresponding \SSM\ for optimal contorl 
can be written as
\begin{align}
\mathbf{f} = \left\{
\begin{aligned}
& x_2 \\
& (\tau_s(x_1,x_2,x_3) - (d(u_3) + b) x_2 ) \inertia^{-1} \\
& x_5 \\
& x_6 \\
& \beta^2( u_1 - x_3) - 2\beta x_5 \\
& \beta^2( u_2 - x_4) - 2\beta x_6
\end{aligned} 
\right.\label{e:SSM_maccepa_}
\end{align}
where $\mathbf{x}=(x_1,x_2,x_3,x_4,x_5,x_6)^\top=(q,\dot{q},\theta_1,\theta_2,\dot{\theta}_1, \dot{\theta}_2)^\top \in \mathbb{R}^6$ is the state vector, $\mathbf{u}=(u_1,u_2,u_3)^\top \in \mathbb{R}^3$ is the control input.

\section*{Appendix B: Dynamic Movement Primitives}\label{app:dmp}
A widely-used formalization is Dynamic Movement Primitive (DMP) proposed by \cite{Schaal2006} and \cite{Ijspeert2002,Ijspeert2013}, based on the idea of modelling movements using dynamical systems. 
Below is a formalization of DMPs for representing actuator variables $\btheta=(\btheta_1,\btheta_2,\btheta_3)^\T$ represents the EP, stiffness motor and damping profile.
\begin{align}
\tau \dot{\btheta} &= \bz \\
\tau \dot{\bz} &= \balpha_z (\bbeta_z ( \bg - \btheta )  -  \bz) + s \bA^\T  \f^{\theta}(s)  \\
\tau \dot{s} &= -\alpha_s s \\
f^{\theta}_{m}(s) &= \frac{\sum_{i=1}^{N}  \psi_i(s) }{\sum_{i=1}^{N}\psi_i(s)} w_{m,i} \label{eq:dmp_forcing}\\
\psi_i(s) &= \exp(-\frac{(s-c_i)^2}{2\sigma_i^2}) \label{eq:dmp_basis}
\end{align}
where $\tau > 0$ represents the duration and $\bg$ is the goal position of $\btheta$, the dynamics of $\btheta$ is regulated by a dynamical system which behaves like a mass-spring-damper model, with gains determined by $\balpha_z,\bbeta_z$. $\f^{\theta}(s)$ is a forcing term manipulating the shape of the trajectory. It is a function in phase variable $s$, whose dynamics makes it asymptotically converge to $0$, in a rate controlled by $\alpha_s$. As a result, The efficacy of forcing term gradually decays to zero. This behaviour is purposely designed in \cite{Ijspeert2002,Ijspeert2013} to enhance convergence of $\btheta$ to the goal $\bg$. In addition to $s$, $\bA$ is added to the forcing term to scale it according to the movement distance, where the $m$-th element $a_m = g_m - \theta_m$ corresponds to $m$-th forcing element $f_{m}^{\theta}$. From \eqref{eq:dmp_forcing} and \eqref{eq:dmp_basis} we can see that the forcing term is defined as the weighted sum of a set of $N$ basis functions, of which each is an exponential function defined by centre point $c_i$ and width factor $\sigma_i$.

\section*{Appendix C: Tracking joint trajectory with inverse dynamics controller}\label{app:ch3_redundancy}
The controller here is general for multiple DOFs. $\bq$ is the joint configuration vector
Suppose that the robot is asked to track a desired trajectory $\{ \bq_{\mathrm{des}},\dot{\bq}_{\mathrm{des}},\ddot{\bq}_{\mathrm{des}} \}$ and satisfies
\begin{align}
(\bqdddot - \bqdddot_\mathrm{des}) + \bK_3 ( \bqddot - \bqddot_\mathrm{des}  ) &+ \bK_2 ( \bqdot - \bqdot_\mathrm{des} ) \nonumber \\
&+ \bK_1 ( \bq - \bq_\mathrm{des} ) = \bO \label{eq:idc_joint_dev1}
\end{align}
Then taking derivatives of
\begin{equation}
\bM(\bq) \bqddot + \bC(\bq,\bqdot) \bqdot + \bG(\bq) = \btau_a(\bq,\btheta)
\end{equation}
yields
\begin{equation}
\bM \bqdddot +  \frac{\partial \bM}{\partial t}\bqddot + \bC \bqddot + \frac{\partial \bC }{ \partial t } \bqdot + \frac{\partial \bG}{\partial t}  = \bJ_{\btheta} \bthetadot + \bJ_\bq \bqdot + \bJ_{\dot{\bq}} \bqddot  \label{eq:idc_joint_dev2}.
\end{equation}

where $\bJ_{\btheta},\bJ_\bq,\bJ_{\dot{\bq}}$ are used to represent the Jacobians of $\btau_a$ w.r.t. $\btheta,\bq,\dot{\bq}$. For the following derivation we assume that $\btheta$ is controlled in velocity domain by $\bv$. 

Combining \eqref{eq:idc_joint_dev1} and \eqref{eq:idc_joint_dev2}, after rearranging them, we get
\begin{align}
\bJ_{\btheta}	 \bv =&   \bM \bqdddot_{\mathrm{des}} - \bJ_\bq \bqdot - \bJ_{\dot{\bq}} \bqddot  + \bC \bqddot \nonumber \\
&+ \frac{\partial \bM}{\partial t}\bqddot + \frac{\partial \bC }{ \partial t} \bqdot + \frac{\partial \bG}{\partial t}  \nonumber \\
+& \bM ( \bK_3 ( \bqddot - \bqddot_\mathrm{des}  ) + \bK_2 ( \bqdot - \bqdot_\mathrm{des} ) + \bK_1 ( \bq - \bq_\mathrm{des} ) ) 
\end{align}
Denote the right-hand side as $\bb$, given a cost metric matrix $\bN$, the control law can be given as
\begin{align}
\bu =& \bN^{-\frac{1}{2}} \pinv{ ( \bJ_{\btheta} \bN^{-\frac{1}{2}} ) } \bb \nonumber \\
&+ \bN^{-\frac{1}{2}}(\bI- \pinv{(\bJ_{\btheta} \bN^{-\frac{1}{2}})} \bJ_{\btheta} \bN^{-\frac{1}{2}} ) \bN^{\frac{1}{2}} \bu_1 \label{eq:idcontroller_torque}
\end{align}
which is a closed form controller with joint feedback.

\end{document}